\documentclass[10pt,twocolumn,letterpaper]{article}
\pdfoutput=1 

\usepackage{cvpr}
\usepackage{times}
\usepackage{epsfig}
\usepackage{graphicx}
\usepackage{amsmath}
\usepackage{amssymb}
\usepackage{xcolor}
\usepackage{caption,subcaption}
\usepackage[normalem]{ulem}
\usepackage[font=small]{caption}
\usepackage[bookmarks=false]{hyperref}

\usepackage{algorithm}
\usepackage[noend]{algpseudocode}
\usepackage{bm}
\definecolor{dgreen}{rgb}{0.0,0.6,0.0} 
\definecolor{dred}{rgb}{0.6,0.0,0.0} 
\definecolor{BrickRed}{rgb}{0.72,0.0,0.0}%
\definecolor{grey}{rgb}{0.6,0.6,0.6}%

\newcommand{\pz}{\phantom{0}}

\cvprfinalcopy 

\begin{document}

\title{
Overcoming Limitations of Mixture Density Networks: \\
A Sampling and Fitting Framework for Multimodal Future Prediction
}

\author{Osama Makansi, Eddy Ilg, \"Ozg\"un \c{C}i\c{c}ek and Thomas Brox\\
University of Freiburg\\
{\tt\small makansio,ilge,cicek,brox@cs.uni-freiburg.de}
}

\maketitle

\begin{abstract}
  Future prediction is a fundamental principle of intelligence that helps plan actions and avoid possible dangers.
  As the future is uncertain to a large extent, modeling the uncertainty and multimodality of the future states is of great relevance. 
  Existing approaches are rather limited in this regard and mostly yield a single hypothesis of the future or, at the best, strongly constrained mixture components that suffer from instabilities in training and mode collapse. 
  In this work, we present an approach that involves the prediction of several samples of the future with a winner-takes-all loss and iterative grouping of samples to multiple modes. 
  Moreover, we discuss how to evaluate predicted multimodal distributions, including the common real scenario, where only a single sample from the ground-truth distribution is available for evaluation. 
  We show on synthetic and real data that the proposed approach triggers good estimates of multimodal distributions and avoids mode collapse. Source code is available at \href{https://github.com/lmb-freiburg/Multimodal-Future-Prediction}{https://github.com/lmb-freiburg/Multimodal-Future-Prediction} 
\end{abstract}

\section{Introduction}

\begin{figure}[t]
\begin{center}
\includegraphics[width=0.6\columnwidth,height=1.7in]{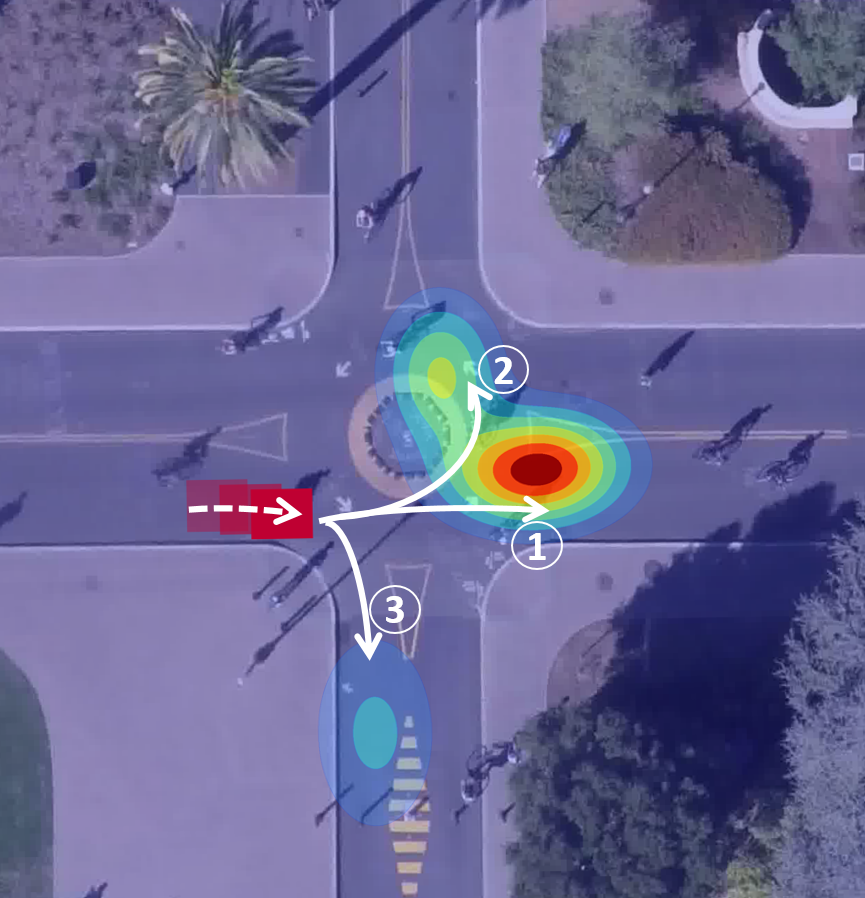} 
\end{center}
  \caption{
    Given the past images, the past positions of an object (red boxes), and the experience from the training data, the approach predicts a multimodal distribution over future states of that object (visualized by the overlaid heatmap). The bicyclist is most likely to move straight (1), but could also continue on the roundabout (2) or turn right (3).
  }
\label{fig:teaser}
\end{figure}

Future prediction at its core is to estimate future states of the environment, given its past states. The more complex the dynamical system of the environment, the more complex the prediction of its future. The future trajectory of a ball in free fall is almost entirely described by deterministic physical laws and can be predicted by a physical formula. If the ball hits a wall, an additional dependency is introduced, which conditions the ball's trajectory on the environment, but it would still be deterministic. 

Outside such restricted physical experiments, future states are typically non-deterministic. Regard the bicycle traffic scenario in Figure~\ref{fig:teaser}. Each bicyclist has a goal where to go, but it is not observable from the outside, thus, making the system non-deterministic. On the other hand, the environment restricts the bicyclists to stay on the lanes and adhere (mostly) to certain traffic rules. Also statistical information on how bicyclists moved in the past in this roundabout and potentially subtle cues like the orientation of the bicycle and its speed can indicate where a bicyclist is more likely to go. A good future prediction must be able to model the multimodality and uncertainty of a non-deterministic system and, at the same time, take all the available conditional information into account to shape the predicted distribution away from a non-informative uniform distribution.   

Existing work on future prediction is mostly restricted to predict a single future state, which often corresponds to the mean of all possible outcomes~\cite{actionanticipate,personloc,streetcrossing,nonparametricuncertainty,trafficactors}. In the best case, such system predicts the most likely of all possible future states, ignoring the other possibilities. As long as the environment stays approximately deterministic, the latter is a viable solution. However, it fails to model other possibilities in a non-deterministic environment, preventing the actor to consider a plan B. 

Rupprecht et al.~\cite{rupprecht} addressed multimodality by predicting diverse hypotheses with the Winner-Takes-All (WTA) loss~\cite{ssvmhyps}, but no distribution and no uncertainty. Conditional Variational Autoencoders (cVAE) provide a way to sample multiple futures~\cite{visualdynamics,bayes,desire}, but also do not yield complete distributions. Many works that predict mixture distributions constrain the mixture components to fixed, pre-defined actions or road lanes ~\cite{mdnhumandriving, mdnsemanticintention}. Optimizing for general, unconstrained mixture distributions requires special initialization and training procedures and suffers from mode collapse; see \cite{rupprecht,cui,curro,messaoud,graves,regMDN}. Their findings are consistent with our experiments. 

In this paper, we present a generic deep learning approach that yields unconstrained multimodal distribution as output and demonstrate its use for future prediction in non-deterministic scenarios. 
In particular, we propose a strategy to avoid the inconsistency problems of the Winner-Takes-All WTA loss, which we name \emph{Evolving WTA} (EWTA). 
Second, we present a two-stage network architecture, where the first stage is based on EWTA, and the second stage fits a distribution to the samples from the first stage. The approach requires only a single forward pass and is simple and efficient. 
In this paper, we apply the approach to future prediction, but it applies to mixture density estimation in general.

To evaluate a predicted multimodal distribution, a ground-truth distribution is required. To this end, we introduce the synthetic Car Pedestrian Interaction (CPI) dataset and evaluate various algorithms on this dataset using the Earth Mover's Distance. In addition, we evaluate on real data, the Standford Drone Dataset (SDD), where ground-truth distributions are not available and the evaluation must be based on a single ground-truth sample of the true distribution. We show that the proposed approach outperforms all baselines. In particular, it prevents mode collapse and leads to more diverse and more accurate distributions than prior work.

\section{Related Work}

\textbf{Classical Future Prediction.}
Future prediction goes back to works like the Kalman filter~\cite{Kalman1960}, linear regression~\cite{McCullagh1989}, autoregressive models~\cite{Walker1925,Yule1927,Akaike1969power}, frequency domain analysis of time series~\cite{Priestly1981}, and Gaussian Processes~\cite{Ohagan1978,Williams1997,Rasmussen2006,Wang2008}. 
These methods are viable baselines, but have problems with high-dimensional data and non-determinism. 

\textbf{Future Prediction with CNNs.}
The possibilities of deep learning have attracted increased interest in future prediction, with examples from various applications: action anticipation from dynamic images~\cite{actionanticipate}, visual path prediction from single image~\cite{visualpathpred}, future semantic segmentation~\cite{futuresemantic}, future person localization~\cite{personloc} and future frame prediction~\cite{futureframe1,futureframe2,futureframe3}. Jin et al.~\cite{sceneparsing} exploited learned motion features to predict scene parsing into the future. Fan et al.~\cite{forecasting} and Luc et al.~\cite{futureinstance} learned feature to feature translation to forecast features into the future. To exploit the time dependency inherent in future prediction, many works use RNNs and LSTMs~\cite{egocentric,unsupervisedlstm,hierachicallmst,longtermlstm,decomposing}. Liu et al.~\cite{ae2} and Rybkin et al.~\cite{ae1} formulated the translation from two consecutive images in a video by an autoencoder to infer the next frame. Jayaraman et al.~\cite{timeagnostic} used a VAE to predict future frames independent of time. 

Due to the uncertain nature of future prediction, many works target predicting uncertainty along with the prediction. Djuric et al.~\cite{trafficactors} predicted the single future trajectories of traffic actors together with their uncertainty as the learned variance of the predictions. Radwan et al.~\cite{streetcrossing} predicted single trajectories of interacting actors along with their uncertainty for the purpose of autonomous street crossing. Ehrhardt et al.~\cite{nonparametricuncertainty} predicted future locations of the objects along with their non-parametric uncertainty maps, which is theoretically not restricted to a single mode. However, it was used and evaluated for a single future outcome. Despite the inherent ambiguity and multimodality in future states, all approaches mentioned above predict only a single future.

\textbf{Multimodal predictions with CNNs.}
Some works proposed methods to obtain multiple solutions from CNNs. 
Guzman-Rivera et al.~\cite{ssvmhyps} introduced the Winner-Takes-All (WTA) loss for SSVMs with multiple hypotheses as output.  
This loss was applied to CNNs for image classification~\cite{ensemblehyps}, semantic segmentation~\cite{ensemblehyps}, image captioning~\cite{ensemblehyps}, and synthesis~\cite{photographichyps}. Firman et al.~\cite{diversenet} used the WTA loss in the presence of multiple ground truth samples. 
The diversity in the hypotheses also motivated Ilg et al.~\cite{uncertainhyps} to use the WTA loss for uncertainty estimation of optical flow. 


Another option is to estimate a complete mixture distribution from a network, like the Mixture Density Networks (MDNs) by Bishop~\cite{mdn}. Prokudin et al.~\cite{directionalgmm} used MDNs with von Mises distributions for pose estimation. Choi et al.~\cite{learnfromdemonstration} utilized MDNs for uncertainties in autonomous driving by using mixture components as samples alternative to dropout~\cite{dropout}. 
However, optimizing for a general mixture distribution comes with problems, such as numerical instability, requirement for good initializations, and collapsing to a single mode~\cite{rupprecht,cui,curro,messaoud,graves,regMDN}.
The Evolving WTA loss and two stage approach proposed in this work addresses these problems.

Some of the above techniques were used for future prediction. 
Vondric et al.~\cite{encoderdecoderk} learned the number of possible actions of objects and humans and the possible outcomes with an encoder-decoder architecture. 
Prediction of a distribution of future states was approached also with conditional variational autoencoders (cVAE). Xue et al.~\cite{visualdynamics} exploited cVAEs for estimating multiple optical flows to be used in future frame synthesis. Lee et al.~\cite{desire} built on cVAEs to predict multiple long-term futures of interacting agents. Li et al.~\cite{flow-grounded} proposed a 3D cVAE for motion encoding.
Bhattacharyya et al.~\cite{bayes} integrated dropout-based Bayesian inference into cVAE. 

The most related work to ours is by Rupprecht et al.~\cite{rupprecht}, where they proposed a relaxed version of WTA (RWTA). They showed that minimizing the RWTA loss is able to capture the possible futures for a car approaching a road crossing, i.e., going straight, turning left, and turning right. Bhattacharyya et al.~\cite{bestofmany} set up this optimization within an LSTM network for future location prediction. Despite capturing the future locations, these works do not provide the whole distribution over the possible locations. 

Few methods predict mixture distributions, but only in a constrained setting, where the number of modes is fixed and the modes are manually bound according to the particular application scenario. Leung et al.~\cite{mdnhumandriving} proposed a recurrent MDN to predict possible driving behaviour constrained to human driving actions on a highway. More recent work by Hu et al.~\cite{mdnsemanticintention} used MDNs to estimate the probability of a car being in another free space in an automated driving scenario. 
In our work, neither the exact number of modes has to be known a priori (only an upper bound is provided), nor does it assume a special problem structure, such as driving lanes in a driving scenario. Another drawback of existing works is that no evaluation for the quality of multimodality is presented other than the performance on the given driving task.

\section{Multimodal Future Prediction Framework\label{sec:mdns}}

\begin{figure}[t]
    
    \begin{subfigure}[t]{0.5\textwidth}
        \includegraphics[width=\textwidth, height=0.9in]{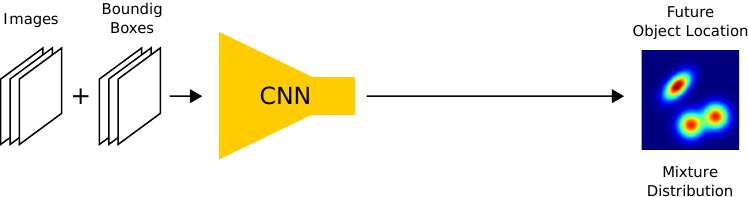}
        \subcaption{
        Direct output of mixture distribution parameters from an encoder.
        \label{fig:mdn}}
    \end{subfigure}
    
    \begin{subfigure}[t]{0.5\textwidth}
        \includegraphics[width=\textwidth, height=0.9in]{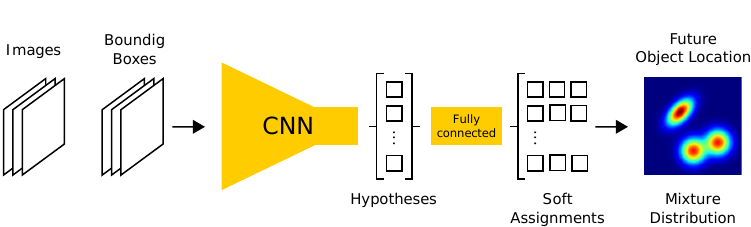}
        \subcaption{
        Our proposed two-stage approach (EWTAD-MDF). The first stage generates hypotheses trained with EWTA loss and the second 
        part fits a mixture distribution by predicting soft assignments of the hypotheses to mixture components. 
        \label{fig:wta_mdn}}
    \end{subfigure}
   \caption{
        Illustration of the normal MDN approach \textbf{(a)} and our proposed extension \textbf{(b)}.
  }  

  \label{fig:schematic}
\end{figure}

Figure~\ref{fig:schematic}b shows a conceptual overview of the approach. The input of the network is the past images and object bounding boxes for the object of interest
$\mathbf{x} = (\mathbf{I}_{t-h},...,\mathbf{I}_{t}, \mathbf{B}_{t-h},...,\mathbf{B}_{t}$), where
$h$ is the length of the history into the past and the bounding boxes $\mathbf{B}_i$ are provided as mask images, where pixels inside the box are $1$ and others are $0$. 
Given $\mathbf{x}$, the goal is to predict a multimodal distribution $p(\mathbf{y}|\mathbf{x})$ of the annotated object's location $\mathbf{y}$ at a fixed time instant $t + \Delta t$ in the future. 

The training data is a set of images, object masks and future ground truth locations: $\mathcal{D}=\{(\mathbf{x}_1, \hat{\mathbf{y}}_1),...,(\mathbf{x}_N, \hat{\mathbf{y}}_N)\}$, where $N$ is the number of samples in the dataset.
Note that this does not provide the ground-truth conditional distribution for $p(\mathbf{y}|\mathbf{x}_i)$, but only a single sample $\mathbf{\hat{y}}_i$ from that distribution.
To have multiple samples of the distribution, the dataset must contain multiple samples with the exact same input $\mathbf{x}_i$, which is very unlikely for high-dimensional inputs. 
The framework is rather supposed to generalize from samples with different input conditions. This makes it an interesting and challenging learning problem, which is self-supervised by nature.  

In general, $p(\mathbf{y}|\mathbf{x})$ can be modeled by a parametric or non-parametric distribution.
The non-parametric distribution can be modeled by a histogram over possible future locations, where each bin corresponds to a pixel. 
A parametric model can be based on a mixture density, such as a mixture of Gaussians. 
In Section~\ref{sec:experiments}, we show that parametric modelling leads to superior results compared to the non-parametric model.

\subsection{MDN Baseline}

A mixture density network (MDN) as in Figure~\ref{fig:schematic}a models the distribution as a mixture of parametric distributions:
\begin{equation}
    p(\mathbf{y}|\mathbf{x}) = \sum_{i=1}^M \pi_i \phi(\mathbf{y}|\bm{\theta}_i) \,\,,
    \label{eq:lmm}
\end{equation}
where $M$ is the number of mixture components, $\phi$ can be any type of parametric distribution with parameters $\theta_i$, and $\pi_i$ is the respective component's weight. In this work, we use Laplace and Gaussian distributions, thus, in the case of the Gaussian, $\theta_i=({\bm{\mu}}_{i}, {\bm{\sigma}}^2_{i})$, with ${\bm{\mu}}_{i}=(\mu_{i,x}, \mu_{i,y})$ being the mean, and $ {\bm{\sigma}}^2_{i} = (\sigma^2_{i,x}, \sigma^2_{i,y})$ the variance of each mixture component. 
We treat x- and y-components as independent, i.e. $\mathcal{\phi}(x,y) = \mathcal{\phi}(x) \cdot \mathcal{\phi}(y)$, because this is usually easier to optimize.  Arbitrary distributions can still be approximated by using multiple mixture components~\cite{mdn}.

The parameters $(\pi_i, \bm{\mu}_{i}, \bm{\sigma}_{i})$ are all outputs of the network and depend on the input data $\mathbf{x}$ (omitted for brevity). When using Laplace distributions for the mixture components, the output becomes the scale parameter $\bm{b_{i}}$ instead of $\bm{\sigma}_{i}$.
For training the network, we minimize the negative log-likelihood (NLL) of \eqref{eq:lmm} \cite{mdn,directionalgmm,mdnhumandriving,learnfromdemonstration,mdnsemanticintention}.

Optimizing all parameters jointly in MDNs is difficult, becomes numerically unstable in higher dimensions, and suffers from degenerate predictions~\cite{rupprecht,cui}. Moreover, MDNs are usually prone to overfitting, which requires special regularization techniques and results in mode collapse~\cite{curro,graves,messaoud,regMDN}.
We use methodology similar to~\cite{regMDN} and sequentially learn first the means, then the variances and finally all parameters jointly. Even though applying such techniques helps training MDNs, the experiments in Section \ref{sec:exp_mixture_density} show that MDNs still suffer from mode collapse.

\subsection{Sampling and Distribution Fitting Framework}

Since direct optimization of MDNs is difficult, we propose to split the problem into sub-tasks: sampling and distribution fitting; see Figure~\ref{fig:wta_mdn}.
The first stage implements the sampling. Motivated by the diversity of hypotheses obtained with the WTA loss~\cite{ssvmhyps,ensemblehyps,photographichyps,rupprecht}, we propose an improved version of this loss and then use it to obtain these samples, which we will keep referring to as \emph{hypotheses} to distinguish them from 
the samples of the training data $\mathcal{D}$.  


Given these hypotheses, one would typically proceed with the EM-algorithm to fit a mixture distribution. Inspired by~\cite{gmmunsupervised}, we rather apply a second network to perform the distribution fitting; see Figure~\ref{fig:wta_mdn}. This yields a faster runtime and the ability to finetune the whole network end-to-end.

\begin{figure*}[t]
    \begin{center}
    \begin{subfigure}[t]{0.465\textwidth}
        \begin{center}
        \includegraphics[width=0.98\textwidth]{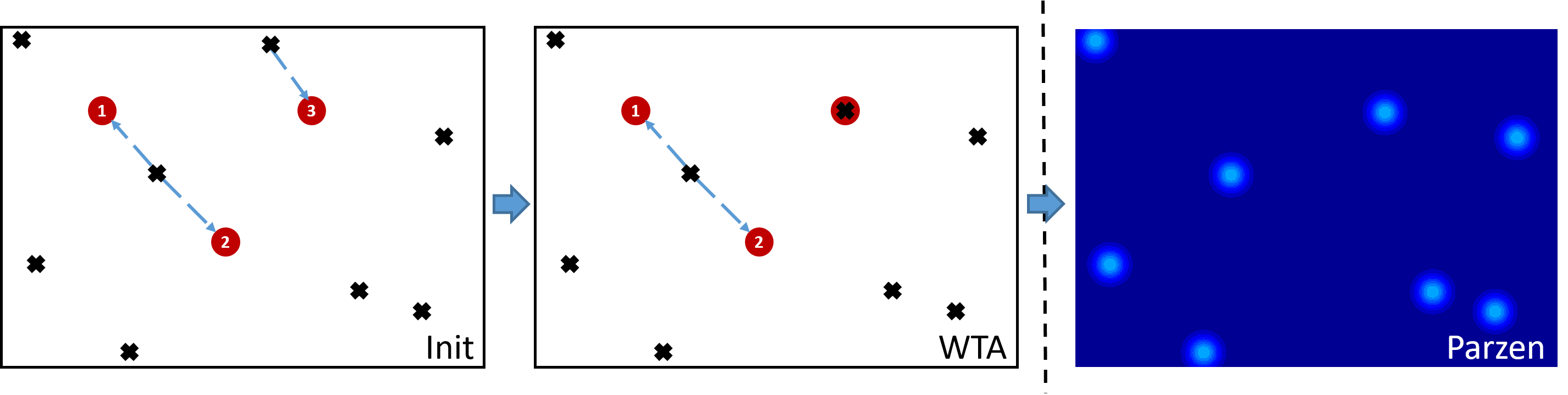}
        \subcaption{
        WTA
        \label{fig:clustering1}}
        \end{center} 
    \end{subfigure}
    \hspace*{1cm}
    \begin{subfigure}[t]{0.465\textwidth}
        \begin{center} 
        \includegraphics[width=0.98\textwidth]{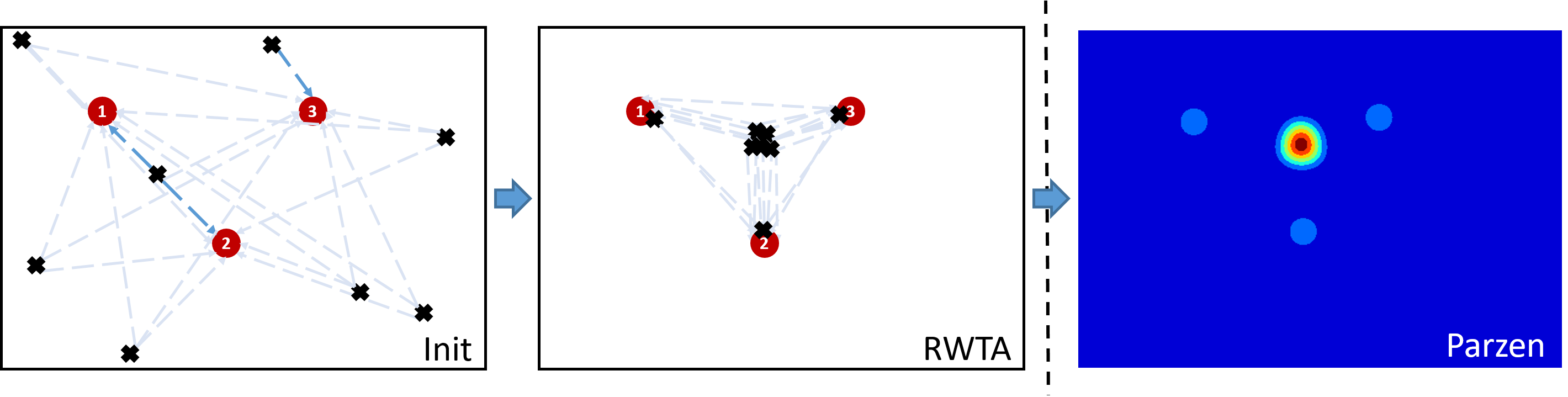}
        \subcaption{
        RWTA
        \label{fig:clustering2}}
        \end{center} 
    \end{subfigure}

    \vspace*{0.3cm}
    \begin{subfigure}[t]{0.94\textwidth}
        \begin{center}
        \includegraphics[width=\textwidth]{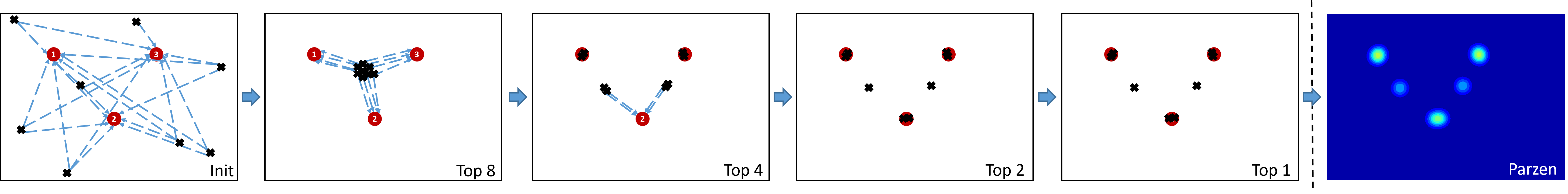}
        \subcaption{
        EWTA
        \label{fig:clustering3}}
        \end{center} 
    \end{subfigure}
   \caption{
        Illustrative example of generating hypotheses with different variants of the WTA loss. Eight hypotheses are generated by the sampling network (crosses) with the purpose to cover the three ground truth samples (numbered red circles). During training, only some ground truth samples are in the minibatch at each iteration. For each, the WTA loss selects the closest hypothesis and the gradient induces an attractive force (indicated by arrows).
        We also show the distributions that arise from applying a Parzen estimator to the final set of hypotheses. 
        \textbf{(a)} In the WTA variant, each ground truth sample selects one winner, resulting in one hypothesis paired with sample 3, one hypothesis in the equilibrium between samples 1 and 2, and the rest never being updated (inconsistent hypotheses). The resulting distribution does not well match the ground truth samples. 
        \textbf{(b)} With the relaxed WTA loss, the non-winning hypotheses are attracted slightly by all samples (thin arrows), moving them slowly to the equilibrium. This increases the chance of single hypotheses to pair with a sample. 
        The resulting distribution contains some probability mass at the ground truth locations, but has a large spurious mode in the center. 
        \textbf{(c)} With the proposed evolving WTA loss, all hypotheses first match with all ground truth samples, moving all hypotheses to the equilibrium (Top 8). Then each ground truth releases 4 hypotheses and pulls only 4 winners, leading to 2 hypotheses pairing with samples 1 and 3 respectively, and 2 hypotheses moving to the equilibrium between samples 1/2 and 2/3, respectively (Top 4). The process continues until each sample selects only one winner (Top 1).
        The resulting distribution has three modes, reflecting the ground truth sample locations well. Only small spurious modes are introduced.}
        \label{fig:wta_variants}
      \end{center}
  
\end{figure*}%

\subsubsection{Sampling - EWTA}
\label{sec:EWTA}

Let $\mathbf{h}_k$ be a hypothesis predicted by our network. We investigate two versions. In the first we model each hypothesis as a point estimate $\mathbf{h}_k = \bm{\mu}_k$ and use the Euclidean distance as a loss function: 
\begin{equation}
l_{ED}(\bm{h}_k, \mathbf{\hat{y}}) =   ||\bm{h}_{k} - \mathbf{\hat{y}}|| \mathrm{\,.} 
\end{equation}
In the second version, we model $\mathbf{h}_k = (\bm{\mu}_k, \bm{\sigma}_k)$ as a unimodal distribution and use the NLL as a loss function~\cite{uncertainhyps}:
\begin{equation}
l_{NLL}(\bm{h}_k, \mathbf{\hat{y}}) =  -\log(\phi(\mathbf{\hat{y}}|\bm{h}_k)) \mathrm{\,.} 
\end{equation}
To obtain diverse hypotheses, we apply the WTA meta-loss~\cite{ssvmhyps,ensemblehyps,photographichyps,rupprecht,uncertainhyps}:
\begin{eqnarray}
    L_{WTA}  &=&  \sum_{k=1}^{K} w_k l(\bm{h}_k, \mathbf{\hat{y}})
    \label{eq:wta1}
        \mathrm{\,,}\\  
    w_i &=& \delta(i=\underset{k}{\mathrm{argmin}}\, ||\bm{\mu}_k - \mathbf{\hat{y}}||)
    \mathrm{\,,}
    \label{eq:wta2}
\end{eqnarray}
where $K$ is the number of estimated hypotheses and $\delta(\cdot)$ is the Kronecker delta, returning $1$ when the condition is true and $0$ otherwise. 
Following~\cite{uncertainhyps}, we always base the winner selection on the Euclidean distance; see \eqref{eq:wta2}. We denote the WTA loss with $l = l_{ED}$ as WTAP (where P stands for Point estimates) and the WTA loss with $l=l_{NLL}$ as WTAD (where D stands for distribution estimates).

Rupprecht et al.~\cite{rupprecht} showed that given a fixed input and multiple ambiguous ground-truth outputs, the WTA loss ideally leads to a Voronoi tessellation of the ground truth. Comparing to the EM-algorithm, this is equivalent to a perfect k-means clustering. However, in practice, k-means is known to depend on the initialization. Moreover, in our case, only one hypothesis is updated at a time (comparable to iterative k-means), the input condition $\mathbf{x}$ is constantly alternating, and we have a CNN in the loop. 

This makes the training process very brittle, as illustrated in Figure~\ref{fig:clustering1}. The red dots here present ground truths, which are iteratively presented one at a time, each time putting a loss on one of the hypotheses (black crosses) and thereby attracting them. 
When the ground truths iterate, it can happen that hypotheses get stuck in an equilibrium (i.e. a hypothesis is attracted by multiple ground truths). In the case of WTA, a ground truth pairs with at most one hypothesis, but one hypothesis can pair with multiple ground truths. In the example from Figure~\ref{fig:clustering1}, this leads to one hypothesis pairing with ground truth 3 and one hypothesis pairing with both, ground truths 1 and 2. This leads to a very bad distribution in the end. 
For details see caption of Figure~\ref{fig:wta_variants}.

Hence, Rupprecht et al.~\cite{rupprecht} relaxed the $\mathrm{argmin}$ operator in \eqref{eq:wta2} and added a small constant $\epsilon$ to all $w_i$ (RWTA), while still ensuring $\sum_i w_i = 1$. 
The effect of the relaxation is illustrated in Figure \ref{fig:clustering2}. 
In comparison to WTA, this results in more hypotheses to pair with ground truths. However, each ground truth also pairs with at most one hypothesis and all excess hypotheses move to the equilibrium. RWTA therefore alleviates the convergence problem of WTA, but still leads to hypotheses generating an artificial, incorrect mode. The resulting distribution also reflects the ground truth samples very badly. This effect is confirmed by our experiments in Section~\ref{sec:experiments}.

We therefore propose another strategy, which we name Evolving WTA (EWTA). In this version, we update the top-$k$ winners. Referring to \eqref{eq:wta2}, this means that $k$ weights are $1$, while $M-k$ weights are $0$. We start with $k=M$ and then decrease $k$ until $k=1$. Whenever $k$ is decreased, a hypothesis previously bound to a ground truth is effectively released from an equilibrium and becomes free to pair with a ground truth. The process is illustrated in Figure \ref{fig:clustering3}. EWTA provides an alternative relaxation, which assures that no residual forces remain. While this still does not guarantee that in odd cases a hypothesis is left in an equilibrium, it leads to much fewer hypotheses being unused than in WTA and RWTA and for a much better distribution of hypotheses in general. The resulting spurious modes are removed later, after adding the second stage and a final end-to-end finetuning of our pipeline.


\subsubsection{Fitting - MDF}

In the second stage of the network, we fit a mixture distribution to the estimated hypotheses (we call this stage Mixture Density Fitting (MDF); see Figure~\ref{fig:wta_mdn}). Similar to Zong et al.~\cite{gmmunsupervised}, we
estimate the soft assignments of each hypothesis to the mixture components:
\begin{eqnarray}
   \bm{\gamma}_k = \mathrm{softmax}(\bm{z}_k) \mathrm{\,,}
   \label{eq:softmax}
\end{eqnarray}
where $k=1..K$ and $\bm{z}_k$ is an $M$-dimensional output vector for each hypothesis $k$. The soft-assignments yield the mixture parameters as follows~\cite{gmmunsupervised}:
\begin{eqnarray}
    \pi_i &=& 
    \frac{1}{K}
    \sum_{k=1}^{K} \gamma_{k,i} \mathrm{\,,}
    \label{eq:pi}\\
    \bm{\mu}_{i} &=& \frac{\sum_{k=1}^{K} \gamma_{k,i} \bm{\mu}_k}{\sum_{k=1}^{K} \gamma_{k,i}} \mathrm{\,,}
   \label{eq:mu}\\
   \bm{\sigma}_{i}^2 &=& \frac{\sum_{k=1}^{K} \gamma_{k,i}\left[(\bm{\mu}_i - \bm{\mu}_{k})^2 + \bm{\sigma}_{k}^2\right]}{\sum_{k=1}^K \gamma_{k,i}} 
   \mathrm{\,.}
    \label{eq:var}
\end{eqnarray}
In Equation \ref{eq:var}, following the law of total variance, we add $\bm{\sigma^2}_k$. This only applies to WTAD. For WTAP $\bm{\sigma^2}_k=0$. 

Finally, we insert the estimated parameters from equations \eqref{eq:pi}, \eqref{eq:mu}, \eqref{eq:var} back into the NLL in \eqref{eq:lmm}. First, we train the two stages of the network sequentially, i.e., we train the fitting network after the sampling network. However, since EWTA does not ensure hypotheses that follow a well-defined distribution in general, we finally remove the EWTA loss and finetune the full network  end-to-end with the NLL loss.

\section{Car Pedestrian Interaction Dataset}
Detailed evaluation of the quality of predicted distributions requires a test set with the ground truth distribution. Such distribution is typically not available for datasets. Especially for real-world datasets, the true underlying distribution is not available, but only one sample from that distribution. Since there exists no future prediction dataset with probabilistic multimodal ground truth, we simulated a dataset based on a static environment and moving objects (a car and a pedestrian) that interact with each other; see Figure~\ref{fig:WTA-multi-stage-qualitative_toy}. The objects move according to defined policies that ensure realistic behaviour and multimodality. Since the policies are known, we can evaluate on the ground-truth distributions $p(\mathbf{y}|\mathbf{x})$ of this dataset. For details we refer to the supplementary material.

\section{Evaluation Metrics}
\textbf{Oracle Error.} For assessing the diversity of the predicted hypotheses, we report the commonly used \textit{Oracle Error}. It is computed by selecting the hypothesis or mode closest to the ground truth. This metric uses the ground truth to select the best from a set of outputs, thus it prefers methods that produce many diverse outputs. Unreasonable outputs are not penalized.


\textbf{NLL.} The Negative Log-Likelihood (NLL) measures the fit of a ground-truth sample to the predicted distribution and allows evaluation on real data, where only a single sample from the ground truth distribution is available. Missing modes and inconsistent modes are both penalized by NLL when being averaged over the whole dataset.
In case of synthetic data with the full ground-truth distribution, we sample from this distribution and average the NLL over all samples. 

\textbf{EMD.} 
If the full ground-truth distribution is available for evaluation, we report the Earth Mover's distance (EMD)~\cite{emd}, also known as Wasserstein metric. As a metric between distributions, it penalizes accurately all differences between the predicted and the ground-truth distribution. One can interpret it as the energy required to move the probability mass of one distribution such that it matches the other distribution, i.e. it considers both, the size of the modes and the distance they must be shifted. The computational complexity of EMD is $O(N^3logN)$ for an $N$-bin histogram and in our case every pixel is a bin. Thus, we use the wavelet approximation WEMD~\cite{wemd}, which has a complexity of $O(N)$. 

\textbf{SEMD.} To make the degree of multimodality of a mixture distribution explicit, we use the EMD to measure the distance between all secondary modes and the primary (MAP) mode, i.e., the EMD to convert a multimodal into a unimodal distribution. We name this metric Self-EMD (SEMD). Large SEMD indicates strong multimodality, while small SEMD indicates unimodality. SEMD is only sensible as a secondary metric besides NLL. 

\section{Experiments\label{sec:experiments}}

\subsection{Training Details}
Our sampling stage is the encoder of the FlowNetS architecture by Dosovitskiy et al.~\cite{flownet} followed by two additional convolutional layers. The fitting stage is composed of two fully connected layers (details in the Supplemental Material).
We choose the first stage to produce $K=40$ hypotheses and the mixture components to be $M=4$.
For the sampling network, we use EWTA and follow a sequential training procedure, i.e., we learn $\bm{\sigma_i}s$ after we learn $\bm{\mu_i}s$. 
We train the sampling and the fitting networks one-by-one. Finally, we remove the EWTA loss and finetune everything end-to-end. The single MDN networks are initialized with the same training procedure as mentioned above before switching to actual training with the NLL loss for a mixture distribution. 

Since the CPI dataset was generated using Gaussian distributions, we use a Gaussian mixture model when training models for the CPI dataset. For the SDD dataset, we choose the Laplace mixture over a Gaussian mixture, because minimizing its negative log-likelihood corresponds to minimizing the L1 distance~\cite{uncertainhyps} and is more robust to outliers.

\subsection{Datasets}
\textbf{CPI Dataset.}
The training part consists of $20$k random samples, while for testing, we randomly pick $54$ samples from the policy. For the time offset into the future we choose $\Delta{t}=20$ frames. We evaluated our method and its baselines on this dataset first, since it allows for quantitative evaluation of distributions. 

\textbf{SDD.}
We use the Stanford Drone Dataset (SDD)~\cite{sdd} to validate our methods on real world data. SDD is composed of drone images taken at the campus of the Stanford University to investigate the rules people follow while navigating and interacting. It includes different classes of traffic actors. We used a split of 50/10 videos for training/testing. For this dataset we set $\Delta{t}=5$ sec. For more details see Supplemental Material.

\newcommand{\rulesep}{\unskip\ \vrule\ }
\begin{figure*}[t]
  \resizebox{\linewidth}{!}{%
  \centering
  \setlength{\tabcolsep}{0.7pt}%
  \begin{tabular}{c|c|c|c|c}
      \includegraphics[width=0.2\textwidth, height=1.1in]{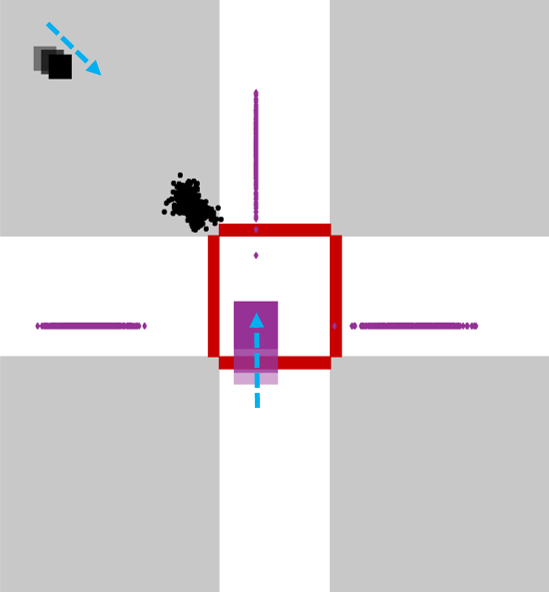}  
      \hspace*{0.5mm}
      & 
      \hspace*{0.5mm}
      \includegraphics[width=0.2\textwidth, height=1.1in]{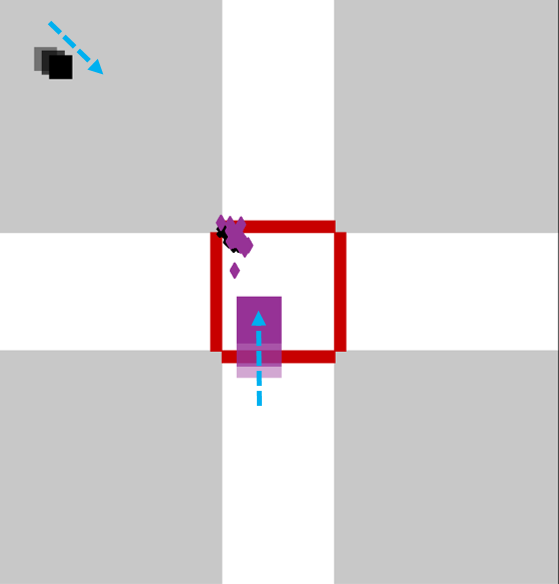}  
      \hspace*{0.5mm}
      & 
      \hspace*{0.5mm}
      \includegraphics[width=0.2\textwidth, height=1.1in]{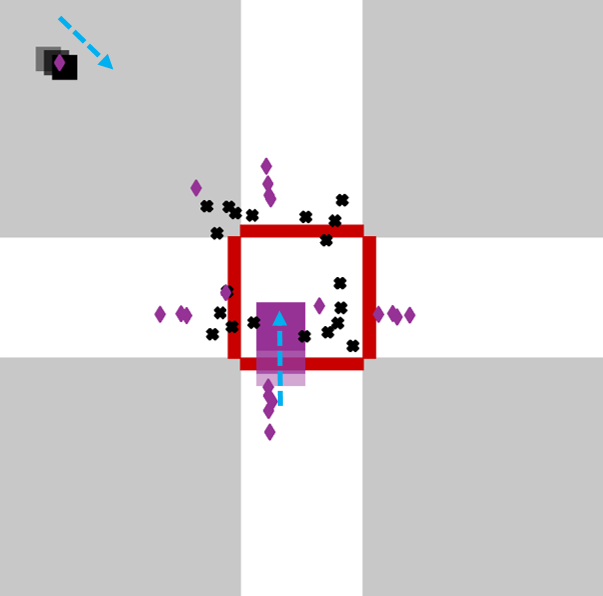}  
      \hspace*{0.5mm}
      & 
      \hspace*{0.5mm}
      \includegraphics[width=0.2\textwidth, height=1.1in]{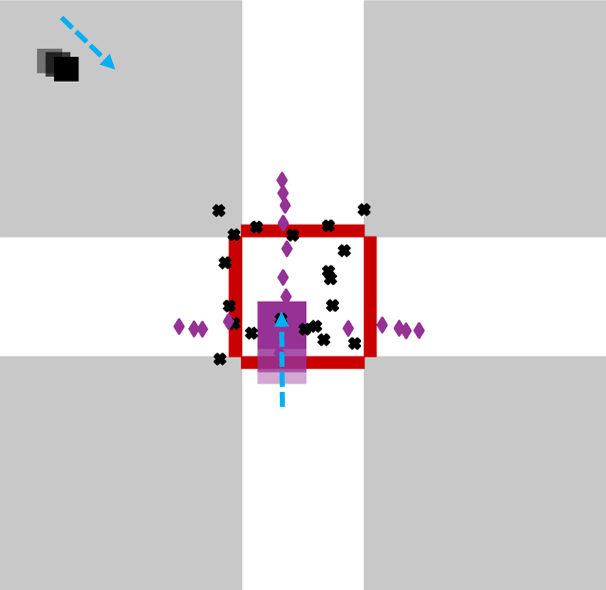}  
      \hspace*{0.5mm}
      &
      \hspace*{0.5mm}
      \includegraphics[width=0.2\textwidth, height=1.1in]{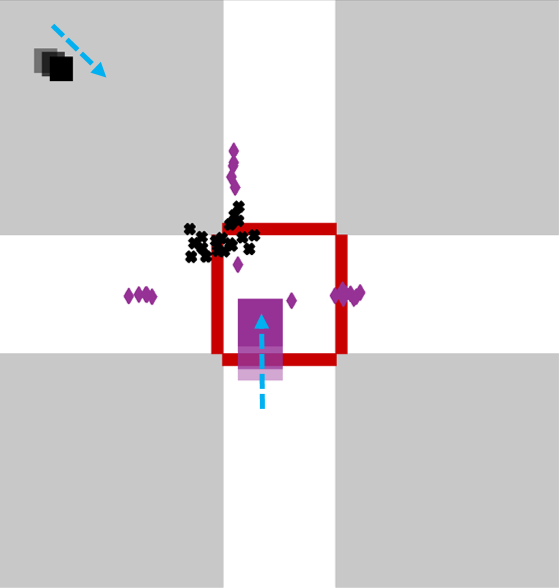}  
      \\%
      GT & Dropout & WTA & Relaxed WTA & Evolving WTA 
      
   \end{tabular}%
   }%
   \caption{Hypotheses generation on the CPI dataset. The dataset has always the same environment of one crossing area (red rectangle) and two objects navigating and interacting (pedestrian and car). In this case, a pedestrian (black rectangle) is heading towards the crossing area (indicated by a blue arrow) and a car (pink rectangle) is entering the crossing area. Left shows the ground-truth distribution for the future locations (after 20 frames) of the pedestrian (black dots) and the car (pink dots). According to the policy to be learned, the pedestrian should wait at the corner until the car passes and the car has three options to exit the crossing. Dropout predicts very similar hypotheses (mode-collapse), while all variants of WTA ensure diversity. The set of hypotheses generated by our evolving WTA additionally approximates the ground-truth distribution.
   } 
    \label{fig:WTA-multi-stage-qualitative_toy}
\end{figure*}

\subsection{Hypotheses prediction}
In our two-staged framework, the fitting stage depends on the quality of the hypotheses. 
To this end, we start with experiments to compare the techniques for hypotheses generation (sampling): WTA, RWTA with $\epsilon$=0.05 and the proposed EWTA. Alternatively one could use dropout~\cite{dropout} to generate multiple hypotheses. Hence, we also compare to this baseline.

The predicted hypotheses can be seen as equal point probability masses and their density leads to a distribution. 
To assess how well the hypotheses reflect the ground-truth distribution of the CPI dataset,
we treat the hypotheses as a uniform mixture of Dirac distributions and compute the EMD between this Dirac mixture and the ground truth. The results in Table~\ref{tab:multi-predictions-toy} show that the proposed EWTA clearly outperforms other variants in terms of EMD, showing that the set of hypotheses from EWTA is better distributed than the sets from RWTA and WTA. WTA and RWTA are better in terms of the oracle error, i.e., the best hypothesis from the set fits a little better than the best hypothesis in EWTA. Clearly, WTA is very well-suited to produce diverse hypotheses, from which one of them will be very good, but it fails on producing hypotheses that represent the samples of the true distribution. This problem is fixed with the proposed Evolving WTA.    

The effect is visualized by the example in Figure~\ref{fig:WTA-multi-stage-qualitative_toy}. The figure also shows that dropout fails to produce diverse hypotheses, which results in a very bad oracle error. Its EMD is better than WTA, but much worse than with the proposed EWTA. 

Figure~\ref{fig:WTA-multi-stage-qualitative_toy} shows that only EWTA and dropout learned the interaction between the car and the pedestrian. WTA provides only the general options for the car (north, east, south and west), and both, WTA and RWTA provide only the general options of the pedestrian to be somewhere on the crossing, regardless of the car. EWTA and dropout learned that the pedestrian should stop, given that the car is entering the crossing. However, dropout fails to estimate the future of the car.

\begin{table}[t]
\centering
\begin{tabular}{|l||c||c|}%
\hline
                             & Oracle Error      & EMD            \\
\hline
Dropout                     & $41.80$      & $3.25$          \\
\hline
WTA~\cite{rupprecht}         & $\textbf{\pz{6.96}}$  & $3.94$          \\
\hline
Relaxed WTA~\cite{rupprecht} & $\pz{7.94}$  & $2.82$          \\
\hline
\hline
Evolving WTA (ours)          & $\pz{9.84}$  & $\textbf{1.89}$ \\
\hline
\end{tabular}
\vspace*{1mm}
    \caption{Comparison between approaches for hypotheses prediction on the CPI dataset.
The overall hypotheses distribution of EWTA matches the ground truth distribution much better, as measured by the Earth Mover's distance (EMD). The high oracle error for Dropout indicates lacking diversity among the hypotheses.  
    }
    \label{tab:multi-predictions-toy} 
    \vspace*{-2mm}
\end{table}

\subsection{Mixture Density Estimation} \label{sec:exp_mixture_density}
We evaluated the distribution prediction with the full network and compare it to several prediction baselines including the standard mixture density network (MDN). Details about the baseline implementations can be found in the supplemental material.

Table~\ref{tab:mixture-model-toy} shows the results for the synthetic CPI dataset, where the full ground-truth distribution is available for evaluation. The results confirm the importance of multimodal predictions. While standard MDNs perform better than single-mode prediction, they frequently suffer from mode collapse, even though they were initialized sequentially with the proposed EWTAP and then EWTAD. The proposed two-stage network avoids this mode collapse and clearly outperforms all other methods. An ablation study between EWTAD-MDF and EWTAP-MDF is given in the supplemental.

\begin{figure*}[t]
  \centering
  \resizebox{0.75\linewidth}{!}{%
  \setlength{\tabcolsep}{0.7pt}%
  \begin{tabular}{ccc}
      \includegraphics[width=0.2\textwidth, height=1.1in]{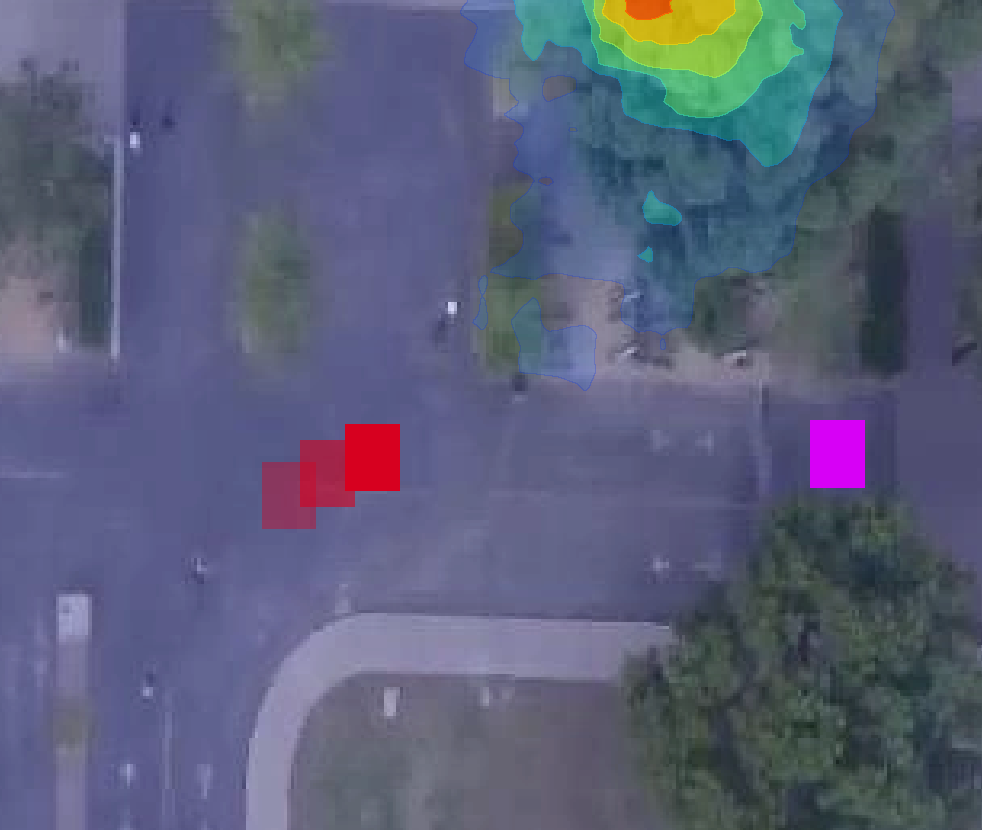}  & 
      \includegraphics[width=0.2\textwidth, height=1.1in]{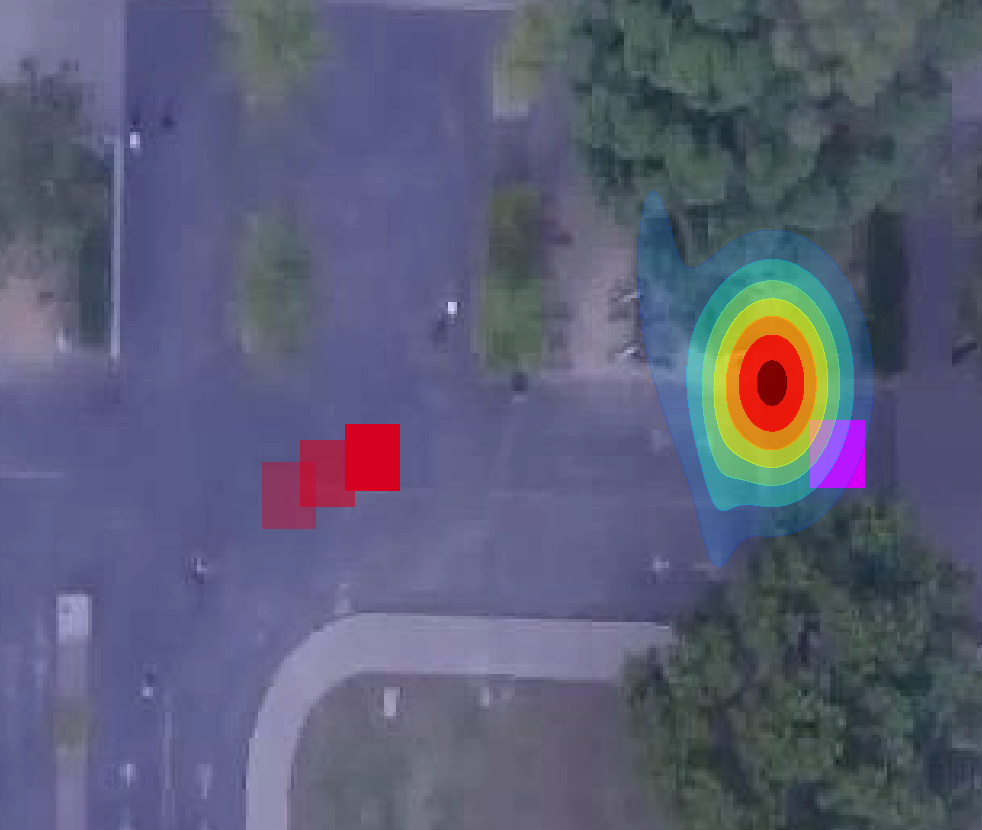}  &
      \includegraphics[width=0.2\textwidth, height=1.1in]{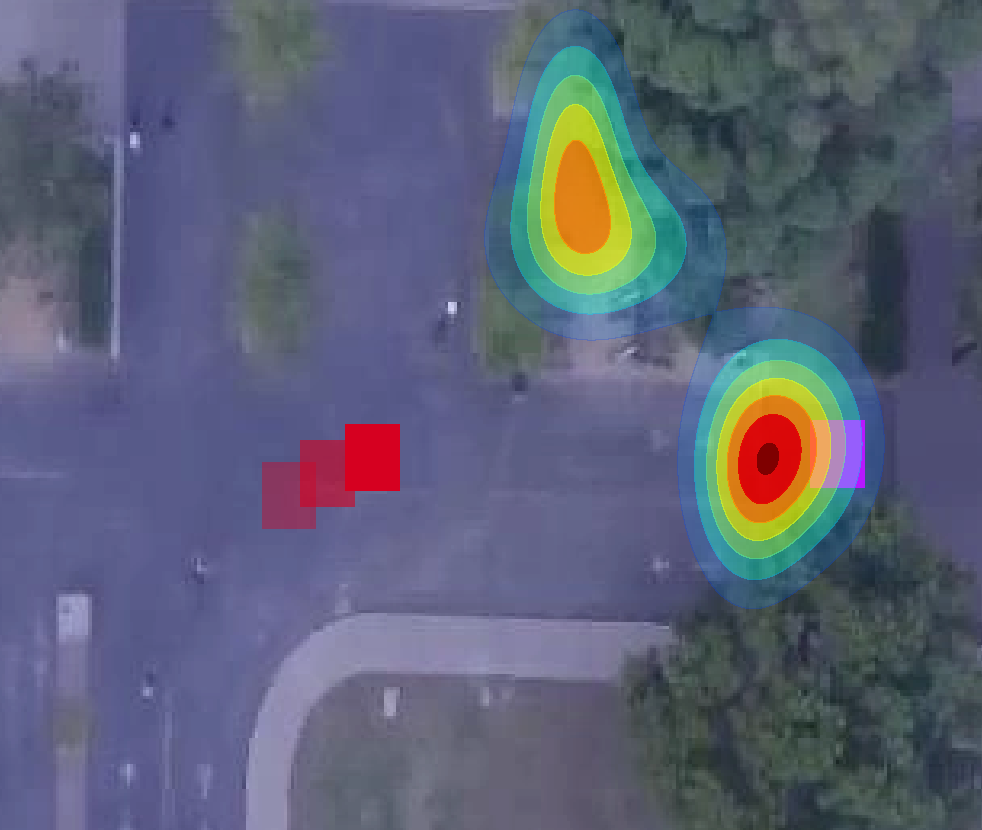}  
      \\
                                                                                    
      \includegraphics[width=0.2\textwidth, height=1.1in]{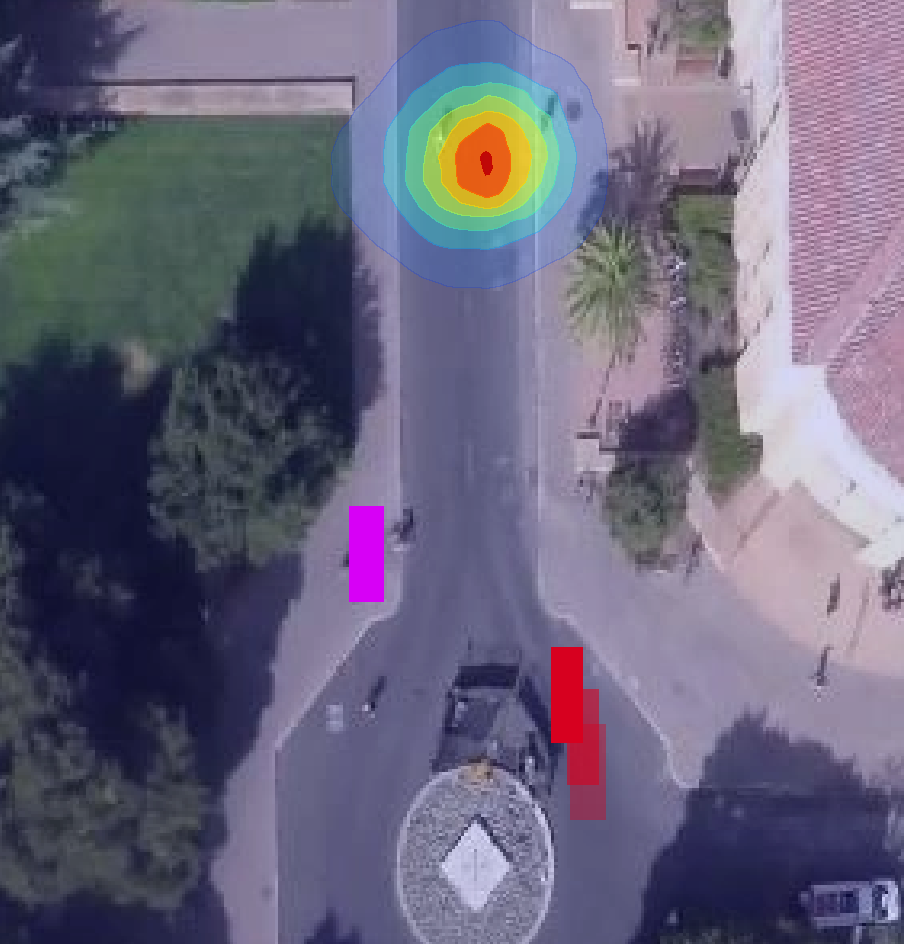}  & 
      \includegraphics[width=0.2\textwidth, height=1.1in]{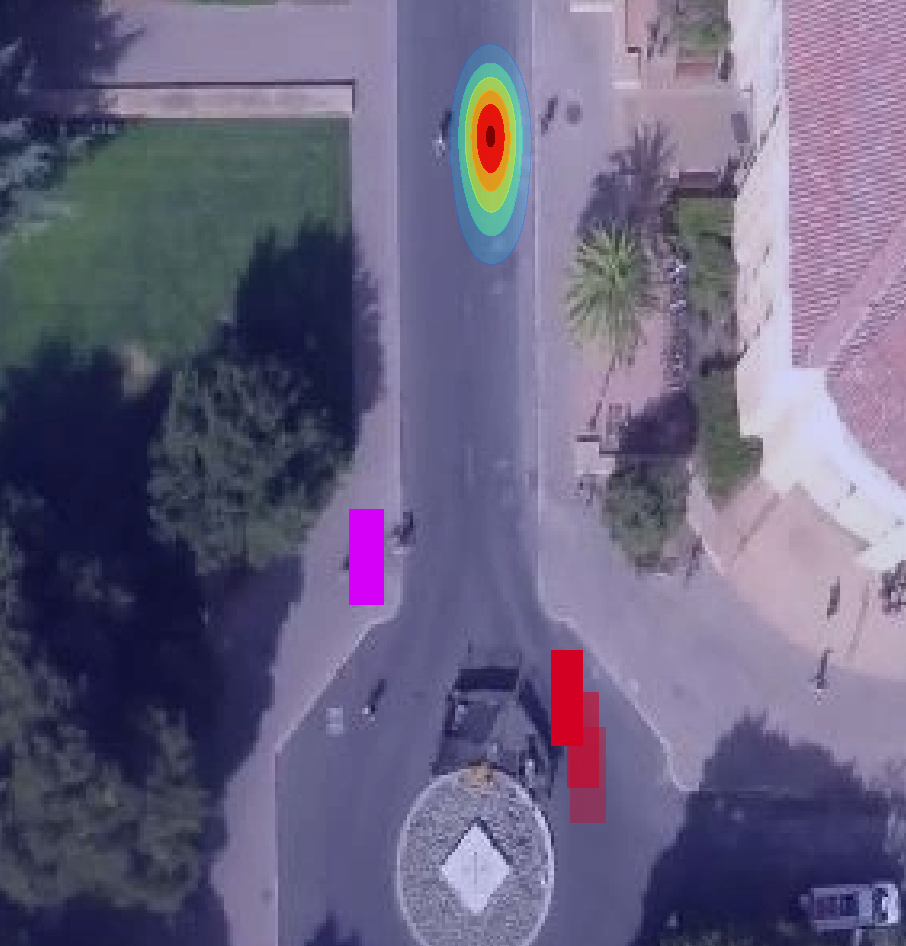}  &
      \includegraphics[width=0.2\textwidth, height=1.1in]{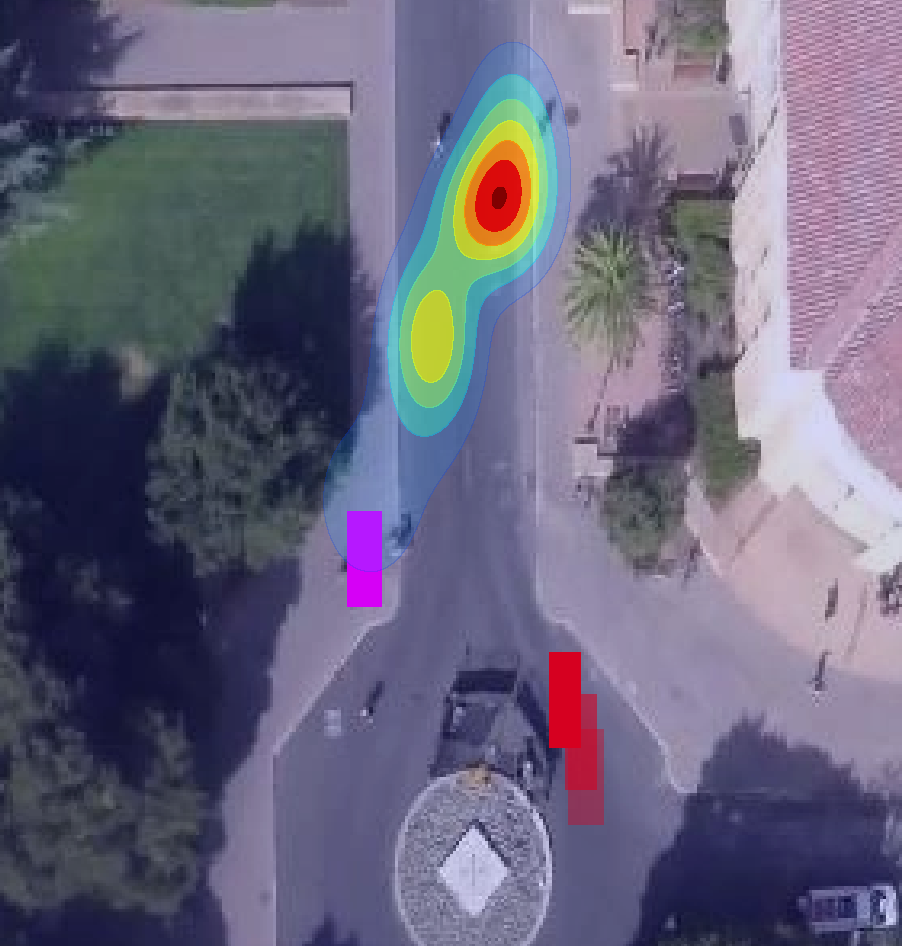}  
      \\
      \includegraphics[width=0.2\textwidth, height=1.1in]{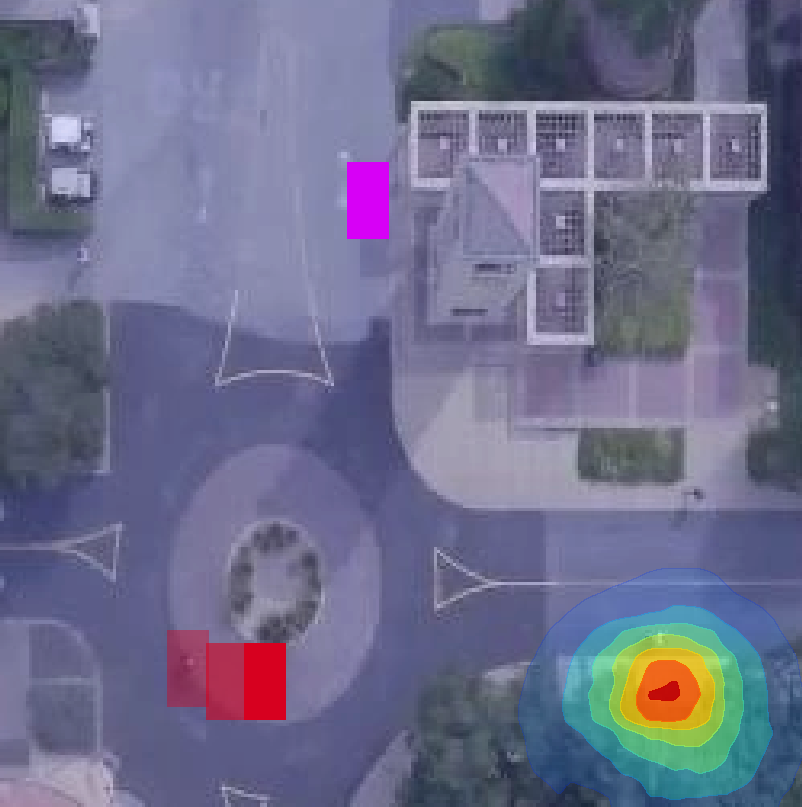}  & 
      \includegraphics[width=0.2\textwidth, height=1.1in]{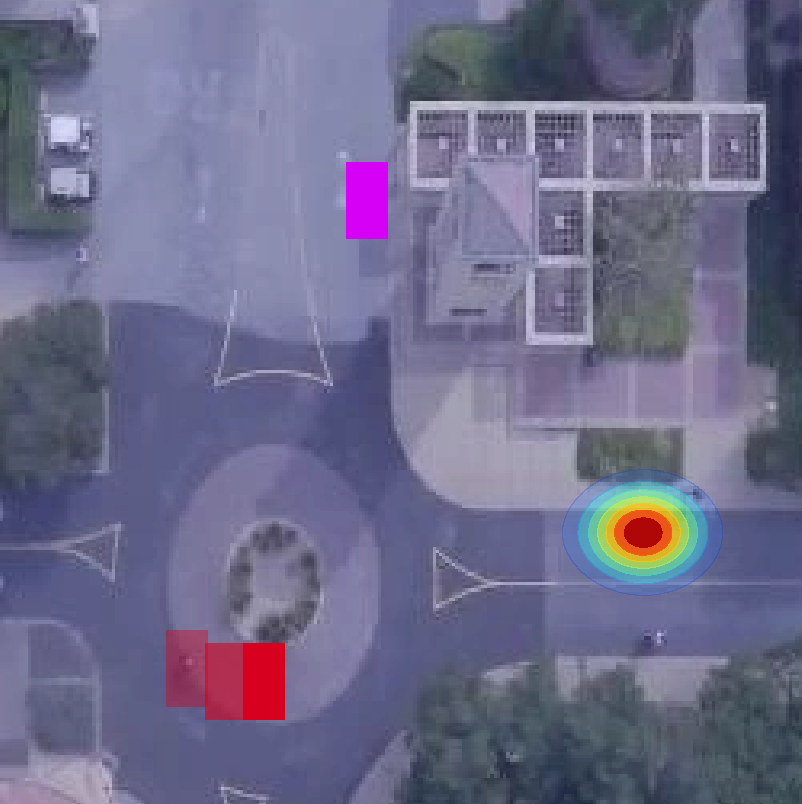}  &
      \includegraphics[width=0.2\textwidth, height=1.1in]{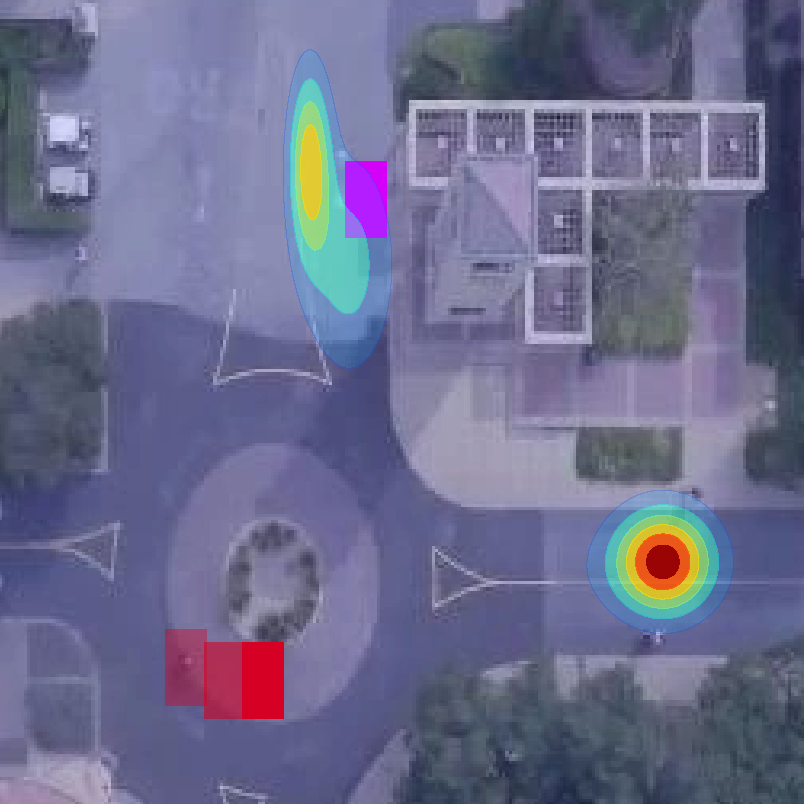}  
      \\
      \small{Non-Parametric} & \small{MDN} & \small{EWTAD-MDF} \\
\end{tabular}%
}%
   \caption{
   Qualitative examples of different multimodal probabilistic methods on SDD. Given three past locations of the target object (red boxes), the task is to predict possible future locations. A heatmap overlay is used to show the predicted distribution over future locations, while the ground truth location is indicated with a magenta box. Both variants of the proposed method capture the multimodality better, while MDN and non-parametric methods reveal overfitting and mode-collapse.} 
    \label{fig:mixture-qualitative}
\end{figure*}

Table~\ref{tab:mixture-model-real} shows the same comparison on the real-world Stanford Drone dataset. Only a single sample from the ground-truth distribution is available here. Thus, we can only compute the NLL and not the EMD. The results confirm the conclusions we obtained for the synthetic dataset: Multimodality is important and the proposed two-stage network outperforms standard MDN. SEMD serves as a measure of multimodality and shows that the proposed approach avoids the problem of mode collapse inherent in MDNs (note that SEMD is only applicable to parametric multimodal distributions).   
This can be observed also in the examples shown in Figure~\ref{fig:mixture-qualitative}.


\begin{table}[t]
\centering
\begin{tabular}{|l||c|c|}%
\hline
Method                   & NLL     & EMD            \\
\hline
Kalman-Filter                 & $25.29$     & $7.03$         \\
\hline
Single Point                  & $-$     & $3.99$         \\
\hline
Unimodal Distribution         & $26.13$ & $2.43$         \\
\hline
\hline
Non-Parametric                & $\pz{9.73}$  & $2.36$          \\
\hline
\hline
MDN                           & $\pz{9.20}$  & $1.83$          \\
\hline
EWTAD-MDF (ours)                    & $\textbf{\pz{8.33}}$  & $\textbf{1.57}$ \\
\hline
\end{tabular}
\vspace*{1mm}
    \caption{Future prediction on the CPI dataset. The results show the importance of multimodality in the prediction model. Classical mixture density networks suffer from frequent mode collapse, which render them inferior to the proposed approach based on EWTA. 
    \label{tab:mixture-model-toy} 
    }
    \vspace*{-2mm}
\end{table}

\begin{table}[t]
\centering
\begin{tabular}{|l||c|c|}%
\hline
Method              & NLL             & SEMD  \\
\hline
Kalman-Filter            & $13.17$         & -      \\
\hline
Unimodal Distribution    & $9.88$          & -      \\
\hline
\hline
Non-Parametric           & $9.35$          & -      \\
\hline
\hline
MDN                      & $9.71$          & $2.36$ \\
\hline
EWTAD-MDF (ours)                & $\textbf{9.33}$ & $\textbf{4.35}$ \\
\hline
\end{tabular}
\vspace*{1mm}
    \caption{Future prediction on the Stanford Drone dataset ($K=20$, $M=4$).
    The two-stage approach yields the best distributions (NLL) 
    and suffers less from mode-collapse than MDN (SEMD).}
    \label{tab:mixture-model-real} 
    \vspace*{-2mm}
\end{table}

In the supplemental material we show more qualitative examples including failure cases and provide ablation studies on some of the design choices. Also a video is provided to show how predictions evolve over time, see \href{https://youtu.be/bIeGpgc2Odc}{https://youtu.be/bIeGpgc2Odc}.

\section{Conclusion}
In this work we contributed to future prediction by addressing the estimation of multimodal distributions. Combining the Winner-Takes-All (WTA) loss for sampling hypotheses and the general principle of mixture density networks (MDNs), we proposed a two-stage sampling and fitting framework that avoids the common mode collapse of MDNs. The major component of this framework is the new way of learning the generation of hypotheses with an evolving strategy. The experiments show that the overall framework can learn interactions between objects and yields very reasonable estimates of multiple possible future states. Although future prediction is a very interesting task, multimodal distribution prediction with deep networks is not restricted to this task. We assume that this work will have impact also in other domains, where distribution estimation plays a role. 

\section{Acknowledgments}
This work was funded in parts by IMRA Europe S.A.S., the German Ministry for Research and Education (BMBF) via the project Deep-PTL and the EU Horizon 2020 project Trimbot 2020.

\clearpage
{\small
\bibliographystyle{ieee}
\bibliography{main}

\begin{thebibliography}{10}\itemsep=-1pt

\bibitem{Yule1927}
Vii. on a method of investigating periodicities disturbed series, with special
  reference to wolfer{\textquoteright}s sunspot numbers.
\newblock {\em Philosophical Transactions of the Royal Society of London A:
  Mathematical, Physical and Engineering Sciences}, 226(636-646):267--298,
  1927.

\bibitem{Akaike1969power}
Hirotugu Akaike.
\newblock Power spectrum estimation through autoregressive model fitting.
\newblock {\em Annals of the institute of Statistical Mathematics},
  21(1):407--419, 1969.

\bibitem{bestofmany}
Apratim Bhattacharyya, Mario Fritz, and Bernt Schiele.
\newblock Accurate and diverse sampling of sequences based on a “best of
  many” sample objective.
\newblock In {\em 31st IEEE Conference on Computer Vision and Pattern
  Recognition}, 2018.

\bibitem{bayes}
Apratim Bhattacharyya, Mario Fritz, and Bernt Schiele.
\newblock Bayesian prediction of future street scenes using synthetic
  likelihoods.
\newblock {\em arXiv preprint arXiv:1810.00746}, 2018.

\bibitem{mdn}
Christopher~M Bishop.
\newblock Mixture density networks.
\newblock Technical report, Citeseer, 1994.

\bibitem{photographichyps}
Qifeng Chen and Vladlen Koltun.
\newblock Photographic image synthesis with cascaded refinement networks.
\newblock In {\em IEEE International Conference on Computer Vision (ICCV)},
  volume~1, page~3, 2017.

\bibitem{learnfromdemonstration}
Sungjoon Choi, Kyungjae Lee, Sungbin Lim, and Songhwai Oh.
\newblock Uncertainty-aware learning from demonstration using mixture density
  networks with sampling-free variance modeling.
\newblock In {\em 2018 IEEE International Conference on Robotics and Automation
  (ICRA)}, pages 6915--6922. IEEE, 2018.

\bibitem{cui}
Henggang Cui, Vladan Radosavljevic, Fang-Chieh Chou, Tsung-Han Lin, Thi Nguyen,
  Tzu-Kuo Huang, Jeff Schneider, and Nemanja Djuric.
\newblock Multimodal trajectory predictions for autonomous driving using deep
  convolutional networks.
\newblock {\em arXiv preprint arXiv:1809.10732}, 2018.

\bibitem{curro}
J. Curro and J. Raquet.
\newblock Deriving confidence from artificial neural networks for navigation.
\newblock In {\em 2018 IEEE/ION Position, Location and Navigation Symposium
  (PLANS)}, pages 1351--1361, April 2018.

\bibitem{trafficactors}
Nemanja Djuric, Vladan Radosavljevic, Henggang Cui, Thi Nguyen, Fang-Chieh
  Chou, Tsung-Han Lin, and Jeff Schneider.
\newblock Motion prediction of traffic actors for autonomous driving using deep
  convolutional networks.
\newblock {\em arXiv preprint arXiv:1808.05819}, 2018.

\bibitem{flownet}
A. Dosovitskiy, P. Fischer, E. Ilg, P. H{\"a}usser, C. Haz{\i}rba{\c{s}}, V.
  Golkov, P. v.d. Smagt, D. Cremers, and T. Brox.
\newblock Flownet: Learning optical flow with convolutional networks.
\newblock In {\em IEEE International Conference on Computer Vision (ICCV)},
  2015.

\bibitem{nonparametricuncertainty}
Sebastien Ehrhardt, Aron Monszpart, Niloy~J Mitra, and Andrea Vedaldi.
\newblock Learning a physical long-term predictor.
\newblock {\em arXiv preprint arXiv:1703.00247}, 2017.

\bibitem{forecasting}
Chenyou Fan, Jangwon Lee, and Michael~S. Ryoo.
\newblock Forecasting hands and objects in future frames, 2017.

\bibitem{diversenet}
Michael Firman, Neill~DF Campbell, Lourdes Agapito, and Gabriel~J Brostow.
\newblock Diversenet: When one right answer is not enough.
\newblock In {\em Proceedings of the IEEE Conference on Computer Vision and
  Pattern Recognition}, pages 5598--5607, 2018.

\bibitem{graves}
Alex Graves.
\newblock Generating sequences with recurrent neural networks.
\newblock {\em CoRR}, abs/1308.0850, 2013.

\bibitem{ssvmhyps}
Abner Guzm\'{a}n-{R}ivera, Dhruv Batra, and Pushmeet Kohli.
\newblock Multiple choice learning: Learning to produce multiple structured
  outputs.
\newblock In F. Pereira, C.~J.~C. Burges, L. Bottou, and K.~Q. Weinberger,
  editors, {\em Advances in Neural Information Processing Systems 25}, pages
  1799--1807. Curran Associates, Inc., 2012.

\bibitem{regMDN}
{Lars U.} Hjorth and {Ian T.} Nabney.
\newblock {\em Regularization of mixture density networks}, volume~2, pages
  521--526.
\newblock Institution of Engineering and Technology (IET), United Kingdom, 470
  edition, 1999.

\bibitem{mdnsemanticintention}
Yeping Hu, Wei Zhan, and Masayoshi Tomizuka.
\newblock Probabilistic prediction of vehicle semantic intention and motion.
\newblock {\em arXiv preprint arXiv:1804.03629}, 2018.

\bibitem{visualpathpred}
S. Huang, X. Li, Z. Zhang, Z. He, F. Wu, W. Liu, J. Tang, and Y. Zhuang.
\newblock Deep learning driven visual path prediction from a single image.
\newblock {\em IEEE Transactions on Image Processing}, 25(12):5892--5904, Dec
  2016.

\bibitem{uncertainhyps}
E. Ilg, {\"O}. {\c{C}}i{\c{c}}ek, S. Galesso, A. Klein, O. Makansi, F. Hutter,
  and T. Brox.
\newblock Uncertainty estimates and multi-hypotheses networks for optical flow.
\newblock In {\em European Conference on Computer Vision (ECCV)}, 2018.
\newblock https://arxiv.org/abs/1802.07095.

\bibitem{timeagnostic}
Dinesh Jayaraman, Frederik Ebert, Alexei~A Efros, and Sergey Levine.
\newblock Time-agnostic prediction: Predicting predictable video frames.
\newblock {\em arXiv preprint arXiv:1808.07784}, 2018.

\bibitem{sceneparsing}
Xiaojie Jin, Huaxin Xiao, Xiaohui Shen, Jimei Yang, Zhe Lin, Yunpeng Chen,
  Zequn Jie, Jiashi Feng, and Shuicheng Yan.
\newblock Predicting scene parsing and motion dynamics in the future.
\newblock In {\em Advances in Neural Information Processing Systems}, pages
  6915--6924, 2017.

\bibitem{Kalman1960}
R.~E. Kalman.
\newblock A new approach to linear filtering and prediction problems.
\newblock {\em ASME Journal of Basic Engineering}, 1960.

\bibitem{desire}
Namhoon Lee, Wongun Choi, Paul Vernaza, Christopher~B Choy, Philip~HS Torr, and
  Manmohan Chandraker.
\newblock Desire: Distant future prediction in dynamic scenes with interacting
  agents.
\newblock In {\em Proceedings of the IEEE Conference on Computer Vision and
  Pattern Recognition}, pages 336--345, 2017.

\bibitem{ensemblehyps}
Stefan Lee, Senthil Purushwalkam~Shiva Prakash, Michael Cogswell, Viresh
  Ranjan, David Crandall, and Dhruv Batra.
\newblock Stochastic multiple choice learning for training diverse deep
  ensembles.
\newblock In {\em Advances in Neural Information Processing Systems}, pages
  2119--2127, 2016.

\bibitem{mdnhumandriving}
K. Leung, E. Schmerling, and M. Pavone.
\newblock Distributional prediction of human driving behaviours using mixture
  density networks.
\newblock Technical report, {Stanford University}, 2016.

\bibitem{flow-grounded}
Yijun Li, Chen Fang, Jimei Yang, Zhaowen Wang, Xin Lu, and Ming-Hsuan Yang.
\newblock Flow-grounded spatial-temporal video prediction from still images.
\newblock In {\em The European Conference on Computer Vision (ECCV)}, September
  2018.

\bibitem{futureframe1}
Wen Liu, Weixin Luo, Dongze Lian, and Shenghua Gao.
\newblock Future frame prediction for anomaly detection – a new baseline.
\newblock In {\em The IEEE Conference on Computer Vision and Pattern
  Recognition (CVPR)}, June 2018.

\bibitem{ae2}
Wenqian Liu, Abhishek Sharma, Octavia Camps, and Mario Sznaier.
\newblock Dyan: A dynamical atoms-based network for video prediction⋆.
\newblock In {\em Proceedings of the European Conference on Computer Vision
  (ECCV)}, pages 170--185, 2018.

\bibitem{futureinstance}
Pauline Luc, Camille Couprie, Yann Lecun, and Jakob Verbeek.
\newblock Predicting future instance segmentations by forecasting convolutional
  features.
\newblock {\em arXiv preprint arXiv:1803.11496}, 2018.

\bibitem{futuresemantic}
Pauline Luc, Natalia Neverova, Camille Couprie, Jacob Verbeek, and Yann LeCun.
\newblock Predicting deeper into the future of semantic segmentation.
\newblock {\em ICCV}, 2017.

\bibitem{Wang2008}
Jack M~Wang, David Fleet, and Aaron Hertzmann.
\newblock Gaussian process dynamical models for human motion.
\newblock 30:283--98, 03 2008.

\bibitem{futureframe3}
Michael Mathieu, Camille Couprie, and Yann LeCun.
\newblock Deep multi-scale video prediction beyond mean square error.
\newblock {\em arXiv preprint arXiv:1511.05440}, 2015.

\bibitem{McCullagh1989}
P. McCullagh and J.~A. Nelder.
\newblock {\em Generalized Linear Models}.
\newblock Chapman \& Hall / CRC, London, 1989.

\bibitem{messaoud}
Safa Messaoud, David Forsyth, and Alexander~G. Schwing.
\newblock Structural consistency and controllability for diverse colorization.
\newblock In Vittorio Ferrari, Martial Hebert, Cristian Sminchisescu, and Yair
  Weiss, editors, {\em Computer Vision -- ECCV 2018}, pages 603--619, Cham,
  2018. Springer International Publishing.

\bibitem{Ohagan1978}
A. O'Hagan and J.~F.~C. Kingman.
\newblock Curve fitting and optimal design for prediction.
\newblock {\em Journal of the Royal Statistical Society. Series B
  (Methodological)}, 40(1):1--42, 1978.

\bibitem{Priestly1981}
M.~B. Priestley.
\newblock {\em {Spectral analysis and time series / M.B. Priestley}}.
\newblock Academic Press, London ; New York :, 1981.

\bibitem{directionalgmm}
Sergey Prokudin, Peter Gehler, and Sebastian Nowozin.
\newblock Deep directional statistics: Pose estimation with uncertainty
  quantification.
\newblock In {\em The European Conference on Computer Vision (ECCV)}, September
  2018.

\bibitem{streetcrossing}
Noha Radwan, Abhinav Valada, and Wolfram Burgard.
\newblock Multimodal interaction-aware motion prediction for autonomous street
  crossing.
\newblock {\em arXiv preprint arXiv:1808.06887}, 2018.

\bibitem{Rasmussen2006}
Carl~Edward Rasmussen.
\newblock Gaussian processes for machine learning.
\newblock MIT Press, 2006.

\bibitem{sdd}
Alexandre Robicquet, Amir Sadeghian, Alexandre Alahi, and Silvio Savarese.
\newblock Learning social etiquette: Human trajectory understanding in crowded
  scenes.
\newblock In {\em European conference on computer vision}, pages 549--565.
  Springer, 2016.

\bibitem{actionanticipate}
Cristian Rodriguez, Basura Fernando, and Hongdong Li.
\newblock Action anticipation by predicting future dynamic images.
\newblock In {\em ECCV'18 workshop on Anticipating Human Behavior}, 2018.

\bibitem{emd}
Y. Rubner, C. Tomasi, and L.~J. Guibas.
\newblock A metric for distributions with applications to image databases.
\newblock In {\em Sixth International Conference on Computer Vision (IEEE Cat.
  No.98CH36271)}, pages 59--66, Jan 1998.

\bibitem{rupprecht}
Christian Rupprecht, Iro Laina, Robert DiPietro, Maximilian Baust, Federico
  Tombari, Nassir Navab, and Gregory~D Hager.
\newblock Learning in an uncertain world: Representing ambiguity through
  multiple hypotheses.
\newblock In {\em International Conference on Computer Vision (ICCV)}, 2017.

\bibitem{ae1}
Oleh Rybkin, Karl Pertsch, Andrew Jaegle, Konstantinos~G Derpanis, and Kostas
  Daniilidis.
\newblock Unsupervised learning of sensorimotor affordances by stochastic
  future prediction.
\newblock {\em arXiv preprint arXiv:1806.09655}, 2018.

\bibitem{wemd}
S. Shirdhonkar and D.~W. Jacobs.
\newblock Approximate earth mover’s distance in linear time.
\newblock In {\em 2008 IEEE Conference on Computer Vision and Pattern
  Recognition}, pages 1--8, June 2008.

\bibitem{dropout}
Nitish Srivastava, Geoffrey Hinton, Alex Krizhevsky, Ilya Sutskever, and Ruslan
  Salakhutdinov.
\newblock Dropout: A simple way to prevent neural networks from overfitting.
\newblock {\em Journal of Machine Learning Research}, 15:1929--1958, 2014.

\bibitem{unsupervisedlstm}
Nitish Srivastava, Elman Mansimov, and Ruslan Salakhudinov.
\newblock Unsupervised learning of video representations using lstms.
\newblock In {\em International conference on machine learning}, pages
  843--852, 2015.

\bibitem{decomposing}
Ruben Villegas, Jimei Yang, Seunghoon Hong, Xunyu Lin, and Honglak Lee.
\newblock Decomposing motion and content for natural video sequence prediction.
\newblock {\em arXiv preprint arXiv:1706.08033}, 2017.

\bibitem{hierachicallmst}
Ruben Villegas, Jimei Yang, Yuliang Zou, Sungryull Sohn, Xunyu Lin, and Honglak
  Lee.
\newblock Learning to generate long-term future via hierarchical prediction.
\newblock {\em arXiv preprint arXiv:1704.05831}, 2017.

\bibitem{encoderdecoderk}
Carl Vondrick, Hamed Pirsiavash, and Antonio Torralba.
\newblock Anticipating visual representations from unlabeled video.
\newblock In {\em Proceedings of the IEEE Conference on Computer Vision and
  Pattern Recognition}, pages 98--106, 2016.

\bibitem{futureframe2}
Vedran Vukotic, Silvia-Laura Pintea, Christian Raymond, Guillaume Gravier, and
  Jan~C Van~Gemert.
\newblock {One-Step Time-Dependent Future Video Frame Prediction with a
  Convolutional Encoder-Decoder Neural Network}.
\newblock In {\em {International Conference of Image Analysis and Processing
  (ICIAP)}}, Proceedings of the 19th International Conference of Image Analysis
  and Processing, Catania, Italy, Sept. 2017.

\bibitem{Walker1925}
Gilbert~T Walker.
\newblock On periodicity.
\newblock {\em Quarterly Journal of the Royal Meteorological Society},
  51(216):337--346, 1925.

\bibitem{longtermlstm}
Nevan Wichers, Ruben Villegas, Dumitru Erhan, and Honglak Lee.
\newblock Hierarchical long-term video prediction without supervision.
\newblock {\em arXiv preprint arXiv:1806.04768}, 2018.

\bibitem{Williams1997}
C.~K.~I. Williams.
\newblock Prediction with gaussian processes: From linear regression to linear
  prediction and beyond.
\newblock In {\em Learning and Inference in Graphical Models}, pages 599--621.
  Kluwer, 1997.

\bibitem{visualdynamics}
Tianfan Xue, Jiajun Wu, Katherine Bouman, and Bill Freeman.
\newblock Visual dynamics: Probabilistic future frame synthesis via cross
  convolutional networks.
\newblock In {\em Advances in Neural Information Processing Systems}, pages
  91--99, 2016.

\bibitem{personloc}
Takuma Yagi, Karttikeya Mangalam, Ryo Yonetani, and Yoichi Sato.
\newblock Future person localization in first-person videos.
\newblock In {\em The IEEE Conference on Computer Vision and Pattern
  Recognition (CVPR)}, June 2018.

\bibitem{egocentric}
Yu Yao, Mingze Xu, Chiho Choi, David~J Crandall, Ella~M Atkins, and Behzad
  Dariush.
\newblock Egocentric vision-based future vehicle localization for intelligent
  driving assistance systems.
\newblock {\em arXiv preprint arXiv:1809.07408}, 2018.

\bibitem{gmmunsupervised}
Bo Zong, Qi Song, Martin~Renqiang Min, Wei Cheng, Cristian Lumezanu, Daeki Cho,
  and Haifeng Chen.
\newblock Deep autoencoding gaussian mixture model for unsupervised anomaly
  detection.
\newblock In {\em International Conference on Learning Representations}, 2018.

\end{thebibliography}
}

\cleardoublepage

\twocolumn[
\null
\vskip .375in
\begin{center}
  {\Large \bf Supplementary Material for: \\ 
Overcoming Limitations of Mixture Density Networks: \\
A Sampling and Fitting Framework for Multimodal Future Prediction \par}
  \vspace*{24pt}
  {
  \large
  \lineskip .5em
  \par
  }
  \vskip .5em
  \vspace*{12pt}
\end{center}]

\setcounter{page}{1}

\setcounter{section}{0} 
\setcounter{figure}{0}
\setcounter{table}{0} 
\setcounter{equation}{0} 

\maketitle

\section{CPI Dataset}

\begin{figure*}[t]
  \centering
  \setlength{\tabcolsep}{0.7pt}%
  \begin{tabular}{c|c|c}
      \includegraphics[width=0.3\textwidth, height=1.7in]{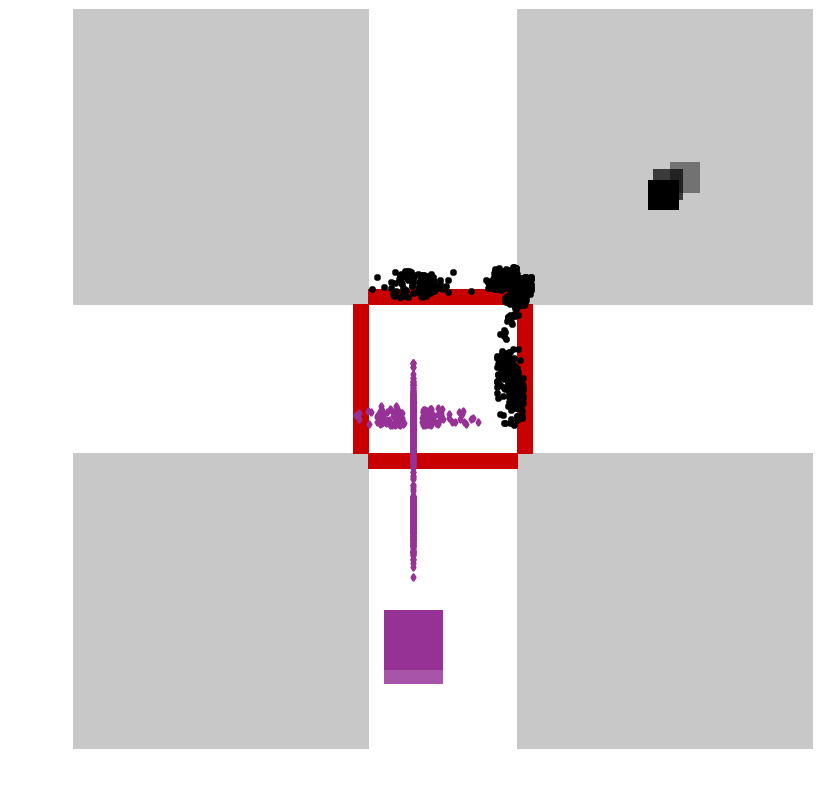}  
      \,\,\,\,\,
      & 
      \includegraphics[width=0.3\textwidth, height=1.7in]{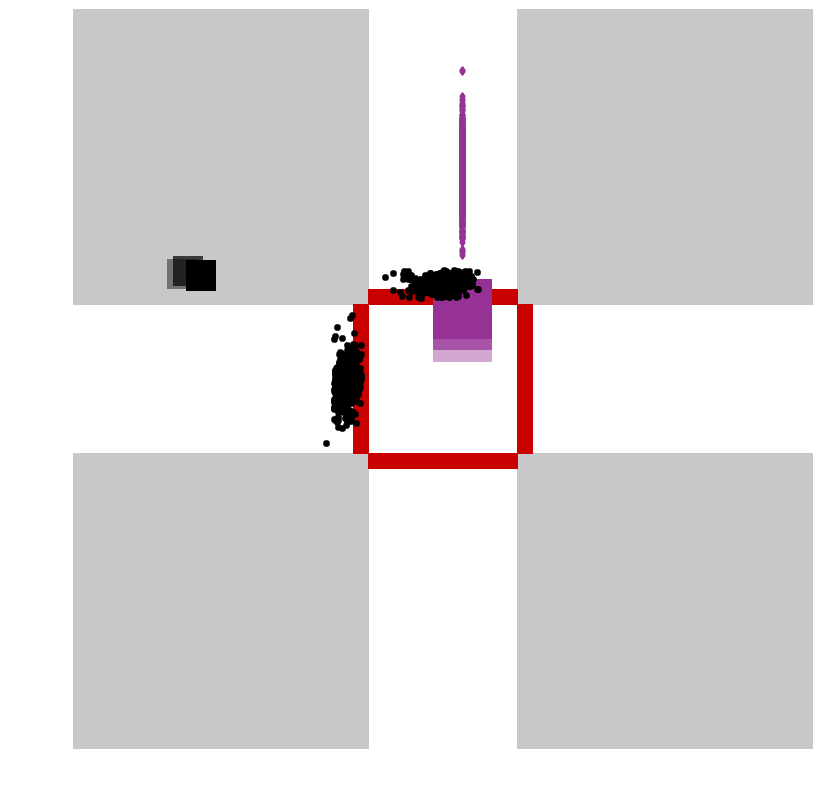}  
      \,\,\,\,\,
      &
      \includegraphics[width=0.3\textwidth, height=1.7in]{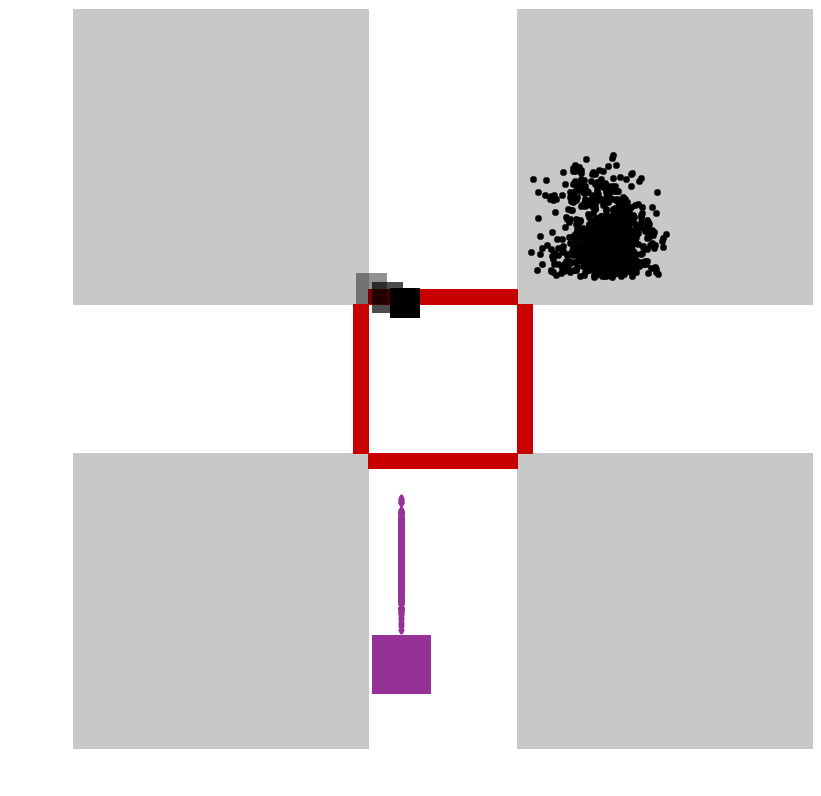}
      \\
      (a) & (b) & (c)
      \\
\end{tabular}%
   \caption{
   Examples from our CPI dataset. Black rectangles denote the current and past locations of the pedestrian, while black dots indicate its future locations ($\Delta t = 20$). Same applies to the car but colored in pink. 
   \textbf{(a)} Pedestrian and car are heading toward the crossing area. The pedestrian must stop at the corner if the car reaches the crossing before, otherwise he can cross over one of the two crossing areas. The car must also stop before the crossing if the pedestrian is crossing or can enter otherwise. 
   \textbf{(b)} The car is leaving the crossing area and therefore only one direction is possible, while the pedestrian does not need to wait and will cross from one of the two possible areas. 
   \textbf{(c)} The pedestrian is in the middle of crossing and the future is unimodal in the destination area. The car needs to wait for the pedestrian to finish crossing. 
   \label{fig:cpi_ex}
} 
\end{figure*}

For evaluating multimodal future predictions, we present a simple toy dataset. 
The dataset consists of a car and a pedestrian and we name it Car Pedestrian Interaction (CPI) dataset.
It is targeted to predicting the future conditioned on this interaction. In the evaluation one can see whether methods just predict independent possible futures for both actors or if they actually constrain these predictions, taking the interactions into account (visible in Figure~4 of the main paper). We show more examples of the data in Figure~\ref{fig:cpi_ex}. The dataset and code to generate it will be made available upon publication. We will now describe the policy used to create the dataset. 

\begin{figure*}[t]
    \centering
  \resizebox{\linewidth}{!}{%
      \centering
      \begin{subfigure}[t]{0.25\textwidth}
        \begin{center}
          \includegraphics[width=\textwidth]{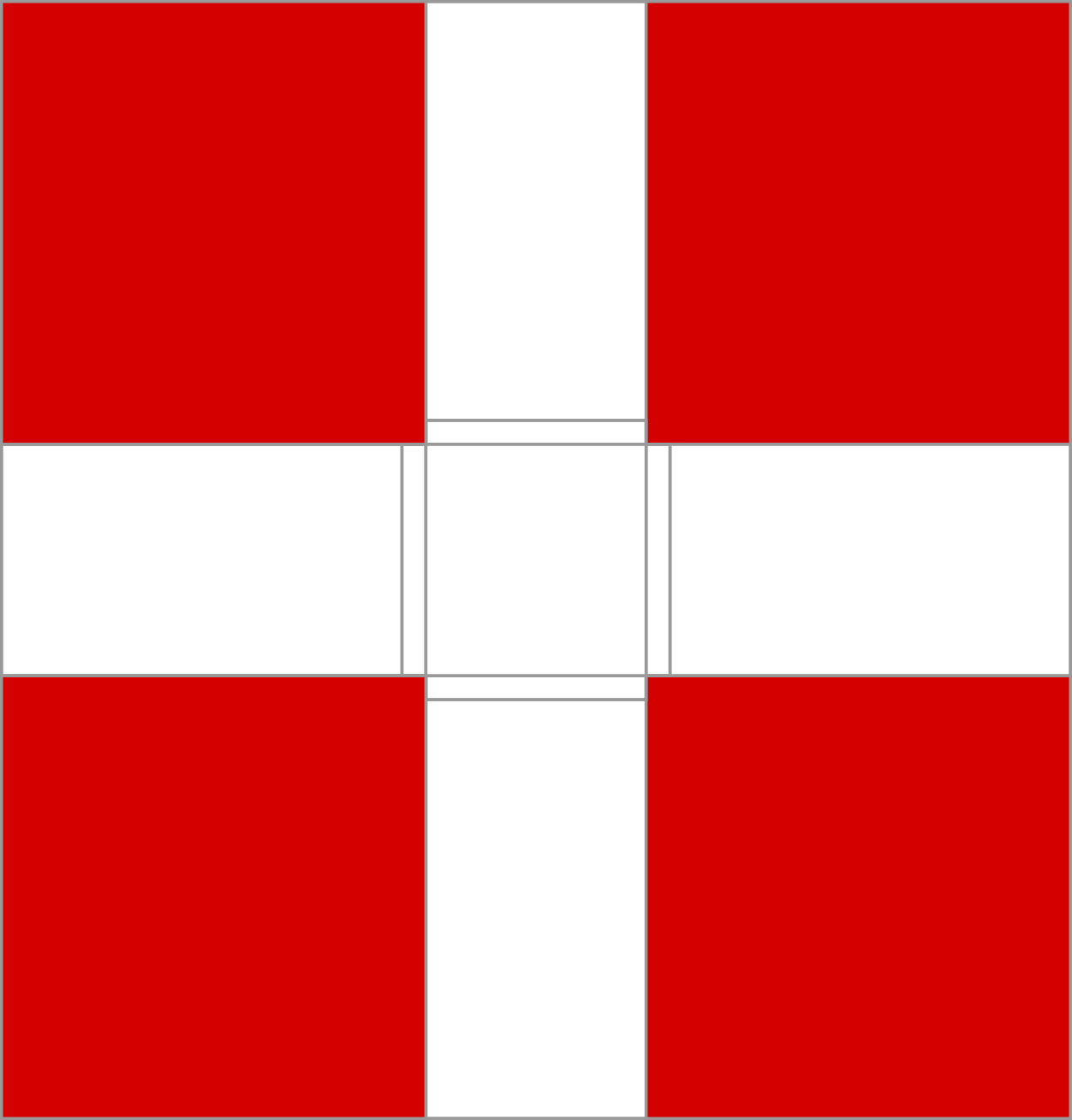}
          \subcaption{Pavement area $R_P$}
        \end{center}
      \end{subfigure}\,\, 
      \begin{subfigure}[t]{0.25\textwidth}
        \begin{center}
          \includegraphics[width=\textwidth]{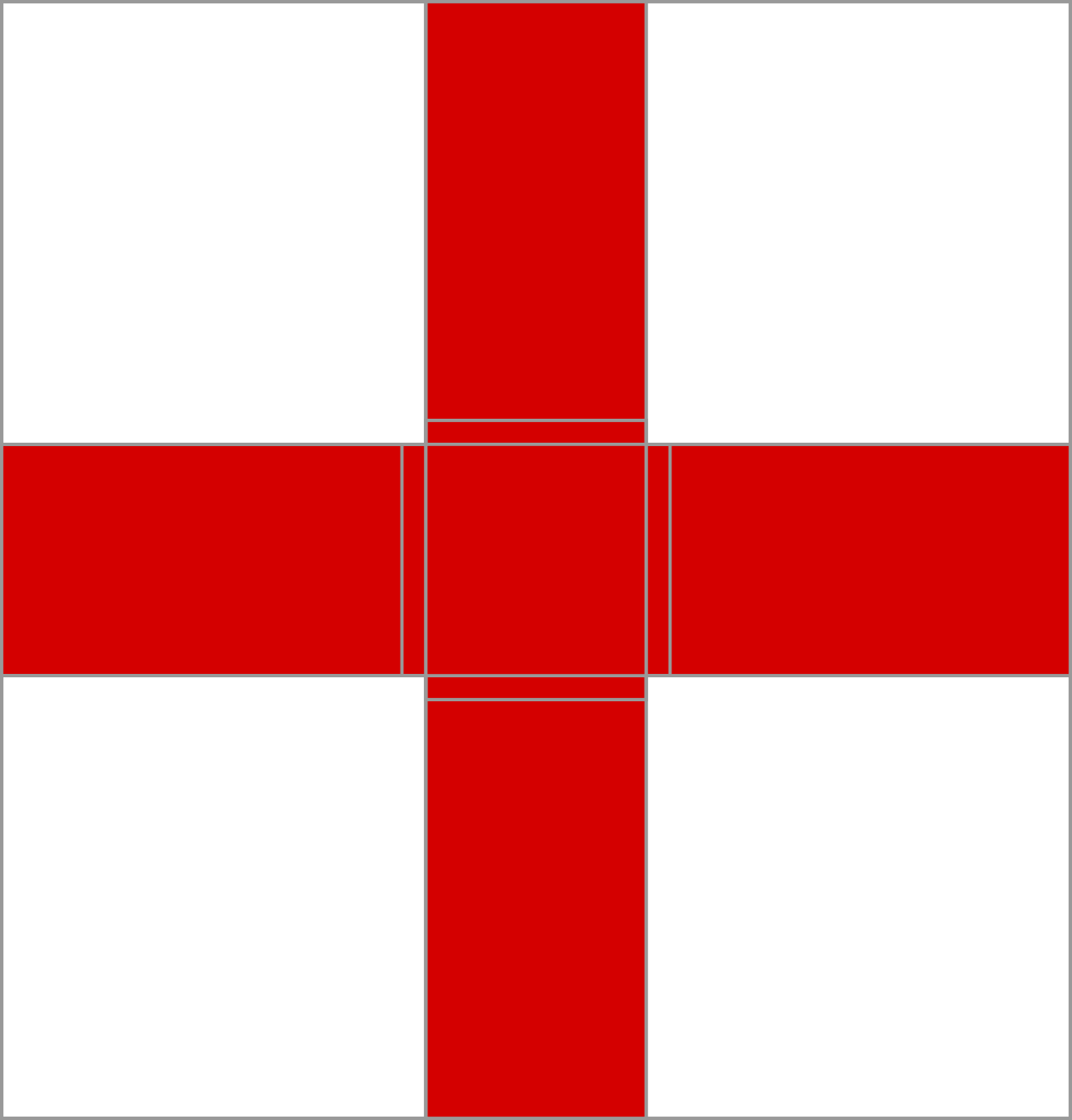}
          \subcaption{Vehicle area $R_V$}
        \end{center}
      \end{subfigure}\,\,
      \begin{subfigure}[t]{0.25\textwidth}
        \begin{center}
          \includegraphics[width=\textwidth]{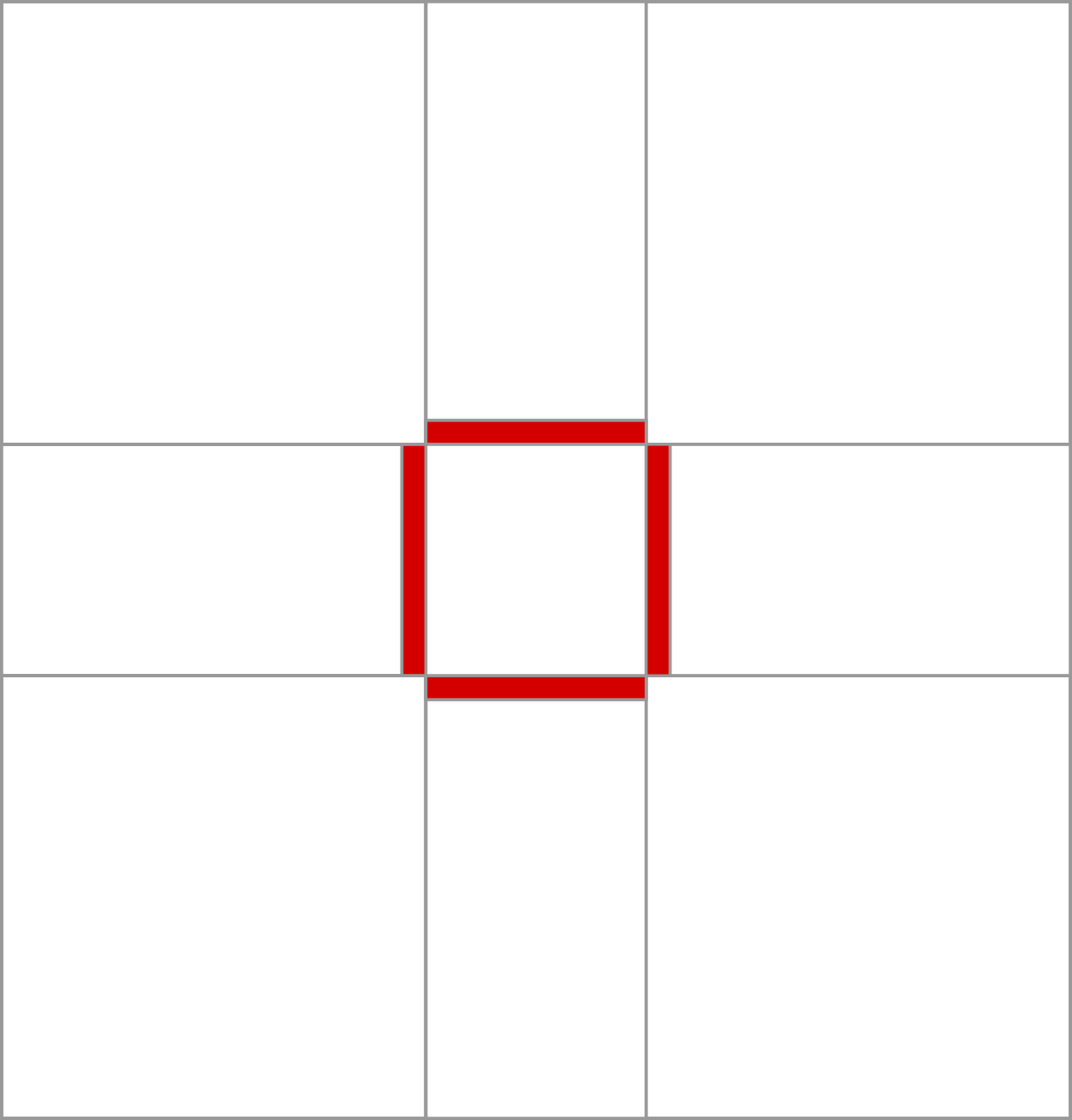}
          \subcaption{Shared area $R_S$}
        \end{center}
      \end{subfigure}\,\,
      \begin{subfigure}[t]{0.25\textwidth}
        \begin{center}
          \includegraphics[width=\textwidth]{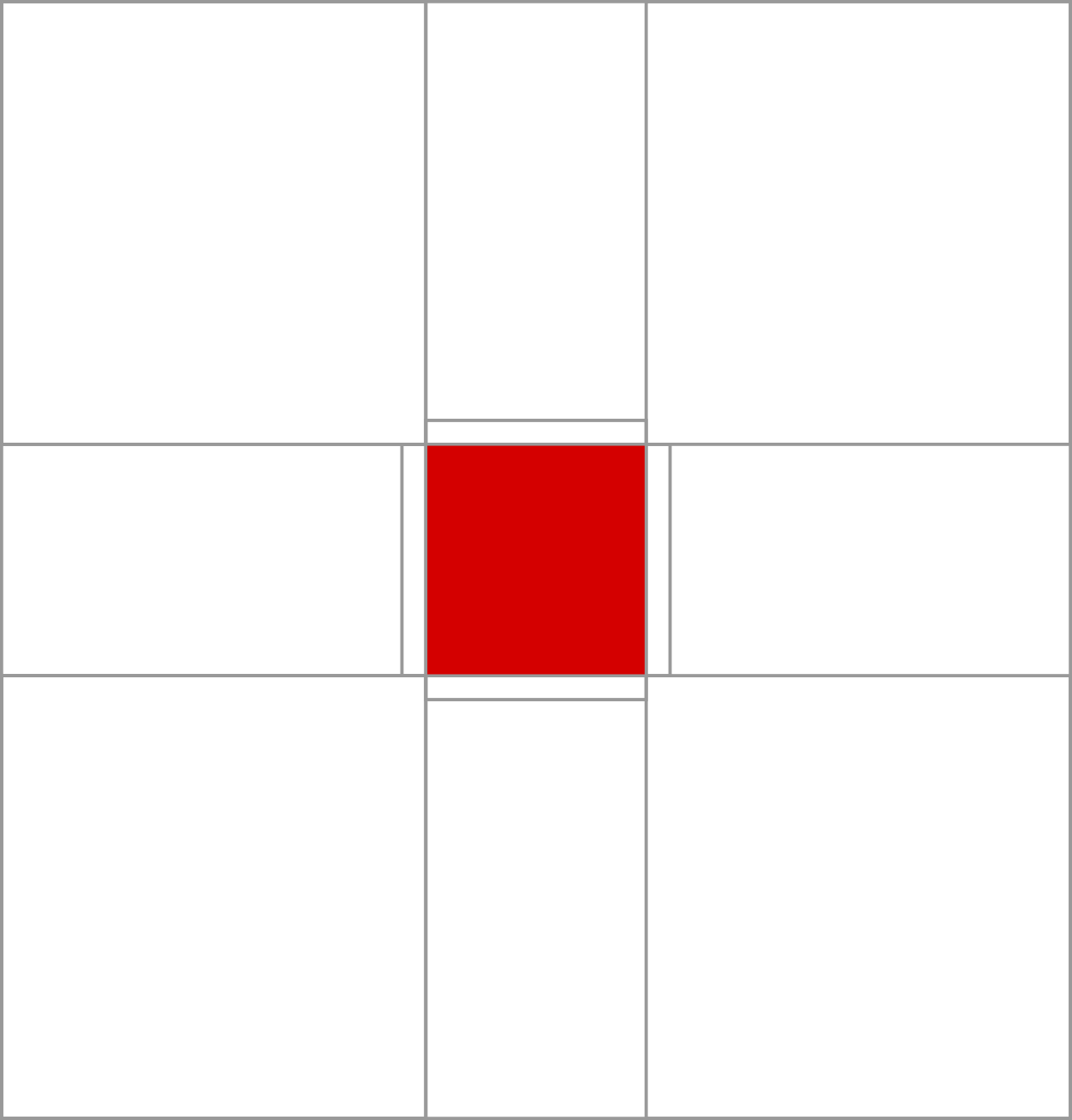}
          \subcaption{Crossing area $R_X$}
        \end{center}
      \end{subfigure}
  }
  \caption{
    Definition of regions of the CPI dataset. 
    \label{fig:area_defs}
  } 
\end{figure*}

Let $\mathbf{x}_{P,t}$, $\mathbf{x}_{C,t}$ denote the locations of car and pedestrian at time $t$.
For the car we define a bounding box of size $40\times40$ pixels and for the pedestrian of size $20\times20$. We denote the pixel regions covered by these boxes by $\mathbf{r}_P$ and $\mathbf{r}_C$ respectively. We furthermore define the areas of the scene shown in Figure~\ref{fig:area_defs}. 
In the beginning of a sequence (for $t=0$), we use rejection sampling to sample valid positions for both actors (such that the pedestrian is contained completely in $R_P \cup R_S$ and the car contained completely in $R_V$).  
We define a sets of possible displacements for pedestrian and car as: 
\begin{eqnarray*}
    \alpha_P &=& \{\mathbf{v}(0^\circ), \mathbf{v}(45^\circ), \mathbf{v}(90^\circ), ..., \mathbf{v}(360^\circ)\}
    \mathrm{\,and}
    \\ 
    \alpha_C &=&  \{\mathbf{v}(0^\circ), \mathbf{v}(90^\circ), \mathbf{v}(180^\circ),  \mathbf{v}(270^\circ)\}
    \mathrm{\,,}
\end{eqnarray*} 
where $\mathbf{v}(\gamma) = 10.0(sin(\gamma), cos(\gamma))$. With adding a displacement to a bounding box $\mathbf{r}$, we indicate that the whole box is shifted. We furthermore define a set of
helper functions given in Table~\ref{table:helper}. 
\begin{table}
    \begin{center}
    \begin{tabular}{|l|l|}
        \hline
        Name & Description \\ 
        \hline
        \hline
        $\mathrm{argmin}_i(x)$ &  $i$-th smallest argument\\         
        \hline
        $\mathrm{argmax}_i(x)$ &  $i$-th largest argument\\         
        \hline
        $\mathrm{dtc}(\mathbf{x})$ & distance of $\mathbf{x}$ to the closest corner\\
        & of the pedestrian area $R_P$  \\
        \hline
        $\mathrm{ad}(\mathbf{a_1}, \mathbf{a}_2)$ & Angle difference between actions $\mathbf{a_1}$ and $\mathbf{a}_2$ \\
        \hline
        $\mathrm{ov}(\mathbf{a}_x, R)$ & Number of pixels overlapping from the \\
        & bounding box $\mathbf{r}_x$ of actor $x$ and region $R$, \\ 
        & after action $\mathbf{a}$ was taken \\
        \hline
    \end{tabular}
    \end{center}
    \caption{
        Helper functions.
        \label{table:helper}
    }
\end{table}
For pedestrian and car, we define the following states:
\begin{eqnarray*}
s_{P,t} &\in& \{\mathrm{TC, SC, C, FC, AC}\} \mathrm{\quad(see\,Table~\ref{table:s_p})\,and} \\
s_{C,t} &\in& \{\mathrm{C, SC, FC, OC}\} \mathrm{\quad(see\,Table~\ref{table:s_c}),}
\end{eqnarray*}
and the world state as: 
\begin{equation*}
\mathbf{w}_t = (\mathbf{x}_{P,t}, \mathbf{x}_{C,t}, E) \mathrm{\,,}
\end{equation*}
where $E$ is the given environment (in this case the crossroad). We define the history of states for pedestrian and car as: 
\begin{eqnarray*}
\mathbf{h}_{P,t} &=& (s_{P,0}, ..., s_{P,t}) \mathrm{\,\,and} \\
\mathbf{h}_{C,t} &=& (s_{C,0}, ..., s_{C,t}) \mathrm{\,\,.}
\end{eqnarray*} 
The current state of pedestrian and car are then determined from their respective histories and the world state: 
\begin{eqnarray*} 
s_{P,t} &=& F_P(\mathbf{h}_{P,t-1}, w_t) \mathrm{\quad(see\,Table~\ref{table:rules_p})\,and} \\ 
s_{C,t} &=& F_C(\mathbf{h}_{C,t-1}, w_t) \mathrm{\quad(see\,Table~\ref{table:rules_c}).} 
\end{eqnarray*}
Given the states, we then define distributions over possible actions and sample from these to update locations:
\begin{eqnarray*}
\mathbf{a}_{P,t} &\sim& \sum_k \pi_k\mathcal{N}(\bm{\mu}_k, \bm{\sigma}_k), \, (\pi_k, \bm{\mu}_k, \bm{\sigma}_k) \in A_P(s_{P,t},s_{C,t})\mathrm{\,,} \\ 
\mathbf{a}_{C,t} &\sim& \sum_k \pi_k\mathcal{N}(\bm{\mu}_k, \bm{\sigma}_k), \, (\pi_k, \bm{\mu}_k, \bm{\sigma}_k) \in A_C(s_{P,t},s_{C,t})\mathrm{\,,} \\ 
\mathbf{x}_{P,t+1} &=& \mathbf{x}_{P,t} + \mathbf{a}_{P,t}\mathrm{\, and}\\ 
\mathbf{x}_{C,t+1} &=& \mathbf{x}_{C,t} + \mathbf{a}_{C,t}\mathrm{\,,}
\end{eqnarray*}
where $A_P(\cdot)$ and $A_C(\cdot)$ are the parameter mapping functions described in Tables~\ref{table:map_p} and~\ref{table:map_c}. We then use this policy to generate $20k$ sequences with three image frames. For each sequence, we generate $10$ different random futures resulting in $200k$ samples for training in total.

\begin{table}
    \begin{center}
    \begin{tabular}{|c|l|}
        \hline
        State & Description \\ 
        \hline
        \hline
        TC & Moving \textbf{T}owards \textbf{C}rossing\\ 
        SC & \textbf{S}tart \textbf{C}rossing\\ 
        C & \textbf{C}rossing\\ 
        FC & \textbf{F}inish \textbf{C}rossing\\ 
        AC & \textbf{A}lready \textbf{C}rossed\\ 
        \hline
    \end{tabular}
    \end{center}
    \caption{
        List of possible pedestrian states. 
        \label{table:s_p}
    }
\end{table}

\begin{table}
    \begin{center}
    \begin{tabular}{|c|l|}
        \hline
        State & Description \\ 
        \hline
        \hline
        SC & \textbf{S}tart \textbf{C}rossing\\ 
        C  & \textbf{C}rossing \\ 
        FC & \textbf{F}inish \textbf{C}rossing\\ 
        OC & \textbf{O}ut of \textbf{C}rossing\\ 
        \hline
    \end{tabular}
    \end{center}
    \caption{
        List of possible car states. 
        \label{table:s_c}
    }
\end{table}

\begin{table}
    \begin{center}
    \begin{tabular}{|c|l|}
        \hline
        \multicolumn{2}{|l|}{$F_P(\mathbf{h}_{P,t-1}, w_t) =$} \\ 
        \hline
        \hline
        TC & \phantom{else }if $\neg$inter$(\mathbf{r}_{P,t}, R_{S})$ and C $\notin \mathbf{h}_{P,t}$ \\
        SC & else if \phantom{$\neg$}inter$(\mathbf{r}_{P,t}, R_{S})$ and C $\notin \mathbf{h}_{P,t}$\\
        FC & else if \phantom{$\neg$}inter$(\mathbf{r}_{P,t}, R_{P})$ and C $\in \mathbf{h}_{P,t}$\\
        C & else if \phantom{$\neg$}inter$(\mathbf{r}_{P,t}, R_{S})$ \\
        AC & else if $\neg$inter$(\mathbf{r}_{P,t}, R_{S})$ and C $\in \mathbf{h}_{P,t}$\\
        \hline
    \end{tabular}
    \end{center}
    \caption{
        List of possible pedestrian states determined by the history and world state. inter$(A,B) = [A \cap B \ne \emptyset]$. For region definitions ($R_S, R_P$) see Figure~\ref{fig:area_defs}.
        \label{table:rules_p}
    }
\end{table}

\begin{table}
    \begin{center}
    \begin{tabular}{|c|l|}
        \hline
        \multicolumn{2}{|l|}{$F_C(\mathbf{h}_{C,t-1}, w_t) =$} \\ 
        \hline
        \hline
        C & \phantom{else} if \phantom{$\neg$}within$(\mathbf{r}_{C,t}, R_{X})$ and C $\notin \mathbf{h}_{C,t}$\\
        SC & else if \phantom{$\neg$}inter$(\mathbf{r}_{C,t}, R_{S})$ and C $\notin \mathbf{h}_{C,t}$\\
        FC & else if \phantom{$\neg$}inter$(\mathbf{r}_{C,t}, R_{S})$ and C $\in \mathbf{h}_{C,t}$\\
        OC & else if $\neg$within$(\mathbf{r}_{C,t}, R_X)$\\
        \hline
    \end{tabular}
    \end{center}
    \caption{
        List of possible pedestrian states determined by the history and world state. inter$(A,B) = [A \cap B \ne \emptyset$]. within$(A,B) = [A \cap B = B]$.  For region definitions ($R_X, R_S$) see Figure~\ref{fig:area_defs}.
        \label{table:rules_c}
    }
\end{table}

\begin{table}
    \begin{center}
    \resizebox{\linewidth}{!}{%
    \begin{tabular}{|c|l|}
        \hline
            $s_{P,t}, s_{C,t}$ & $\{(\pi_1, \bm{\mu}_1, \bm{\sigma}_1), ..., (\pi_n, \bm{\mu}_n, \bm{\sigma}_n)\} = A_P(s_{P,t}, s_{C,t})$ \\
        \hline
        \hline
            TC,* 
            &  $\mu_1 = \underset{\mathbf{a\in\alpha_P}}{\mathrm{argmin}_1}(\mathrm{dtc}(\mathbf{x}_{P,t}+\mathbf{a}))$ \\
            &  $\mu_2 = \underset{\mathbf{a\in\alpha_P}}{\mathrm{argmin}_2}(\mathrm{dtc}(\mathbf{x}_{P,t}+\mathbf{a}))$ \\
            &  $\pi = (0.7, 0.3)$  \\              
            &  $\sigma_i = 2.0$  \\
        \hline 
            SC,\{SC,C,FC\} 
            &  $\mu_1 = (0,0)$ \\
            &  $\pi_1 = 1.0$  \\              
            &  $\sigma_1 = 0$  \\
        \hline 
            SC,OC
            &  $\mu_1 = \underset{\mathbf{a\in\alpha_P}}{\mathrm{argmin}_1}(\mathrm{ad}(\mathbf{a}, \mathbf{a}_{P,t-1}) - \mathrm{ov}(\mathbf{a}, R_S))$ \\
            &  $\pi_1 = 1.0$  \\              
            &  $\sigma_1 = 2.0$  \\
        \hline 
            C,*
            &  $\mu_1 = \underset{\mathbf{a\in\alpha_P}}{\mathrm{argmin}_1}(2*\mathrm{ad}(\mathbf{a}, \mathbf{a}_{P,t-1}) - \mathrm{ov}(\mathbf{a}, R_S))$ \\
            &  $\pi_1 = 1.0$  \\              
            &  $\sigma_1 = 2.0$  \\
        \hline 
            FC,*
            &  $\mu_1 = \underset{\mathbf{a\in\alpha_P}}{\mathrm{argmin}_1}(\mathrm{ad}(\mathbf{a}, \mathbf{a}_{P,t-1}) - \mathrm{ov}(\mathbf{a}, R_P))$ \\
            &  $\pi_1 = 1.0$  \\              
            &  $\sigma_1 = 2.0$  \\
        \hline
            AC,*
            &  $\mu_i = \underset{\mathbf{a\in\alpha_P}}{\mathrm{argmax}_i}(\mathrm{dtc}(\mathbf{x}_{P,t}+\mathbf{a})) $ for $ i = 1..4$ \\
            &  $\pi = (0.4, 0.2, 0.2, 0.2)$ \\
            &  $\sigma_i = 2.0$  \\
        \hline
    \end{tabular}
    }
    \end{center}
    \caption{
        State to distribution parameter mapping for the pedestrian. 
        \label{table:map_p}
    }
\end{table}

\begin{table}
    \begin{center}
    \resizebox{\linewidth}{!}{%
    \begin{tabular}{|c|l|}
        \hline
            $s_{P,t}, s_{C,t}$ & $\{(\pi_1, \bm{\mu}_1, \bm{\sigma}_1), ..., (\pi_n, \bm{\mu}_n, \bm{\sigma}_n)\} = A_C(s_{P,t}, s_{C,t})$ \\
        \hline
        \hline
            \{C, SC, FC\}, *
            &  $\mu_1 = (0,0)$ \\
            &  $\pi_1 = 1.0$  \\              
            &  $\sigma_1 = 0$  \\
        \hline
            *, C
            &  $\mu_i = \underset{\mathbf{a\in\alpha_C}}{\mathrm{argmin}_i}(\mathrm{ad}(\mathbf{a}, \mathbf{a}_{C,t-1})) $ for $ i=1..3$\\
            &  $\pi_i = 1/3$  \\              
            &  $\sigma_i = 2.0$  \\
        \hline
            *, \{FC,SC\}
            &  $\mu_1 = \underset{\mathbf{a\in\alpha_C}}{\mathrm{argmin}_1}(\mathrm{ad}(\mathbf{a}, \mathbf{a}_{C,t-1}))$ \\
            &  $\mu_2 = (0,0)$ \\
            &  $\pi = (0.7, 0.3)$  \\              
            &  $\sigma = (2.0, 0)$  \\
        \hline
            *, \{OC\}
            &  $\mu_1 = \underset{\mathbf{a\in\alpha_C}}{\mathrm{argmin}_1}(\mathrm{ad}(\mathbf{a}, \mathbf{a}_{C,t-1}))$ \\
            &  $\mu_2 = (0,0)$ \\
            &  $\pi = (0.8, 0.3)$  \\              
            &  $\sigma = (2.0, 0)$  \\
        \hline
    \end{tabular}
    }
    \end{center}
    \caption{
        State to distribution parameter mapping for the car. 
        \label{table:map_c}
    }
\end{table}


\pagebreak
\section{Architecture}

We base our architecture on the encoder of FlowNetS~\cite{flownet}. Architecture details are given in Table~\ref{table:archs}.

\begin{table}
   \begin{center}
  \resizebox{0.98\linewidth}{!}{%
  \begin{tabular}{|l|c|cc|c|}
      \hline
      Name & Ch I/O & InpRes & OutRes & Input\\
      \hline
      \hline
      fc7 & 1024/1024 & $8\times8$ & $1\times1$ & conv6a \\
      fc8 & 1024/1024 & $1\times1$ & $1\times1$ & fc7 \\
      fc9 & 1024/$N1$ & $1\times1$ & $1\times1$ & fc8 \\
      \hline
      \hline
      fc10 & $N_{CH1}$/500 & $1\times1$ & $1\times1$ & fc9 \\
      droup-out & 500/500 & $1\times1$ & $1\times1$ & fc10 \\
      fc11 & 500/$N2$ & $1\times1$ & $1\times1$ & drop-out \\
      \hline
    \end{tabular}
   }
  \end{center}
  \caption{
  The top part oSupplementary Material for: \\ 
Overcoming Limitations of Mixture Density Networks: \\
A Sampling and Fitting Framework for Multimodal Future Predictionf the table indicates our base architecture used for MDNs and our first stage. Outputs $N1$ depend on the number of possible output parameters. 
  The bottom part shows the proposed the Mixture Density Fitting (MDF) stage. 
  Outputs $N2$ depend on the number of possible output parameters. 
  Drop-out is performed with dropping probability of $0.5$.
  \label{table:archs}
  }
\end{table}

\pagebreak
\section{Baselines}
\subsection{Kalman Filter}
The Kalman filter is a linear filter for time series observations, which contains process and observation noise~\cite{Kalman1960}. It aims to get better estimates of a dynamic process. It is applied recursively. At each time step there are two phases: predict and update. 

In the predict phase, the future prediction for $t+1$ is calculated given the previous prediction at $t$. For this purpose, a model of the underlying process needs to be defined. We define our process over the vector $\mathbf{x}$ of $(\mathrm{location}, \mathrm{velocity})$ and uncertainties \textbf{P}. The equations integrating the predictions are then: \begin{eqnarray*}
    \textbf{x}_{t+1}' &=& \textbf{F}\cdot{\textbf{x}_{t}}\mathrm{\,,} \\
    \textbf{P}_{t+1}' &=& \textbf{F}\cdot{\textbf{P}_{t}'}\cdot{\textbf{F}^T} + \textbf{Q}\mathrm{\,,}
\end{eqnarray*}
where $\textbf{F}$ is defined as the matrix $(1, \Delta_{t}; 0, 1)$ and \textbf{Q} is the process noise. We do not assume any control from outside and assume constant motion. We compute this constant motion as the average of 2 velocities we get from our history of locations.

In the update phase, the future prediction is computed using the observation $\textbf{z}_{t+1}$ as follows:
\begin{eqnarray*}
    \textbf{K} &=& \textbf{P}_{t+1}'\cdot{({\textbf{P}_{t+1}'} + \textbf{R}_{t+1})^{-1}} \mathrm{\,,} \\
    \textbf{x}_{t+1} &=& \textbf{x}_{t+1}' + \textbf{K}\cdot{(\textbf{z}_{t+1} - \textbf{x}_{t+1}'}) \mathrm{\,,} \\
    \textbf{P}_{t+1} &=& \textbf{P}_{t+1}' - \textbf{K}\cdot{\textbf{P}_{t+1}'} \mathrm{\,,}
\end{eqnarray*}
where \textbf{R} is the observation noise.

For our task we can iterate predict and update only $3$ times, since we are given $2$ history and $1$ current observation. However, since our task is future prediction at $t+\Delta t$ and we assume to not have any more observations until (and including) the last time point, we perform the predict phase at the last iteration $k$ times with the constant motion we assumed. This can be seen as extrapolation by constant motion on top of Kalman filtered observations. In this manner the Kalman filter is a robust linear extrapolation to the future with an additional uncertainty estimate. In our experiments the process and the observation noises are both set to $2.0$.

\begin{table}
   \begin{center}
  \begin{tabular}{|c|c|}
      \hline
      $\sigma_{NP}$ & EMD \\
      \hline
      \,1.0 & $2.39$ \\ 
      \,3.0 & $\textbf{2.35}$ \\ 
      \,4.0 & $3.32$ \\ 
      10.0 & $5.07$ \\ 
      \hline
    \end{tabular}
  \end{center}
  \caption{
  Comparison study on the kernel width of the non-parametric baseline. 
  \label{table:kernelwidth}
  }
  
\end{table}

\subsection{Single Point}
For the single point prediction, we apply the first stage of the architecture from Table~\ref{table:archs}, but we only output a single future position. We train this using the Euclidean Distance loss $l_{ED}$ (Equation (2) of the main paper). 

\subsection{Distribution Prediction}
For the distribution prediction, we apply the first stage of the architecture from Table~\ref{table:archs}, but we output only mean and variance for a unimodal future distribution. We train this using the NLL loss $l_{NLL}$ (Equation (3) of the main paper). 

\subsection{Non-parametric}
In this variant we use the FlowNetS architecture~\cite{flownet}. 
The possible future locations are discretized into pixels and a 
probability $q_\mathbf{y}$ for each pixel $\mathbf{y}$ is output through a softmax from the encoder/decoder network. 

This transforms the problem into a classification problem, for which a one-hot encoding is usually used as ground truth, assigning a probability of $1$ to the true location and $0$ to all other locations. However, in this case such an encoding is much too peaked and would only update a single pixel. In practice we therefore blur the one-hot encoding by a Gaussian with variance $\sigma_{NP}$ (also referred to as soft-classification~\cite{gcnet}).

We then minimize the cross-entropy between the output $q_\mathbf{y}$ and the distribution $\phi(\mathbf{y}|\mathbf{\hat{y}},\sigma_{NP})$ (proportional to the KL-Divergence): 
\begin{equation*} 
L_{NP}=\mathrm{min} [- \sum_{\mathbf{y}} \phi(\mathbf{y}|\mathbf{\hat{y}},\sigma_{NP})\log (q_\mathbf{y})] \mathrm{\,.}
\end{equation*}
We try three different values for $\sigma_{NP}$ as shown in Table~\ref{table:kernelwidth} and use $\sigma_{NP}=3.0$ in practice. 

\section{Training Details}

\begin{figure*}[t]
  \begin{center}
   \begin{subfigure}[t]{0.45\linewidth}
     \centering
      \includegraphics[width=\linewidth]{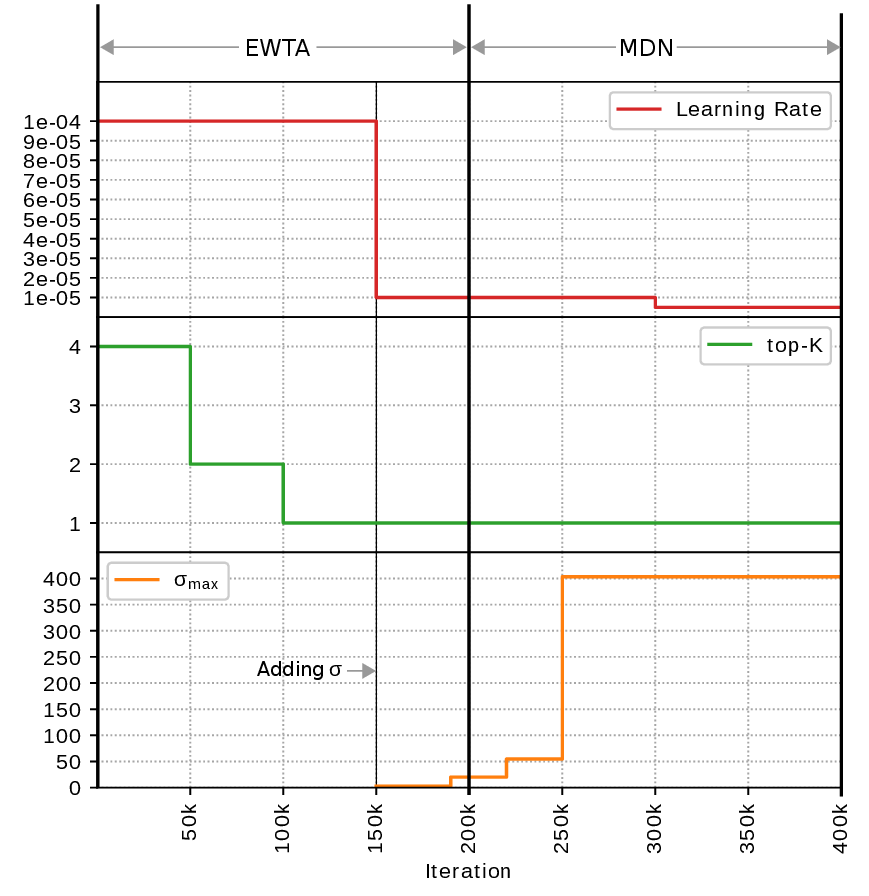}
      \caption{\label{fig:training_mdn}
      Training schedule for MDNs. 
      We first train for 150k iterations using EWTA, optimizing only the means with $l_{ED}$ (Equation 2 of main paper). At 150k, we switch the loss and optimize $l_{NLL}$ (Equation 3 of the main paper) to obtain also variances. 
      At 200k we switch from EWTA to the full mixture density NLL loss. To stabilize the training we set an upper bound on $\sigma$, which we increase during training. 
      }
   \end{subfigure}\,\,
   \begin{subfigure}[t]{0.45\linewidth}
      \centering
      \includegraphics[width=\linewidth]{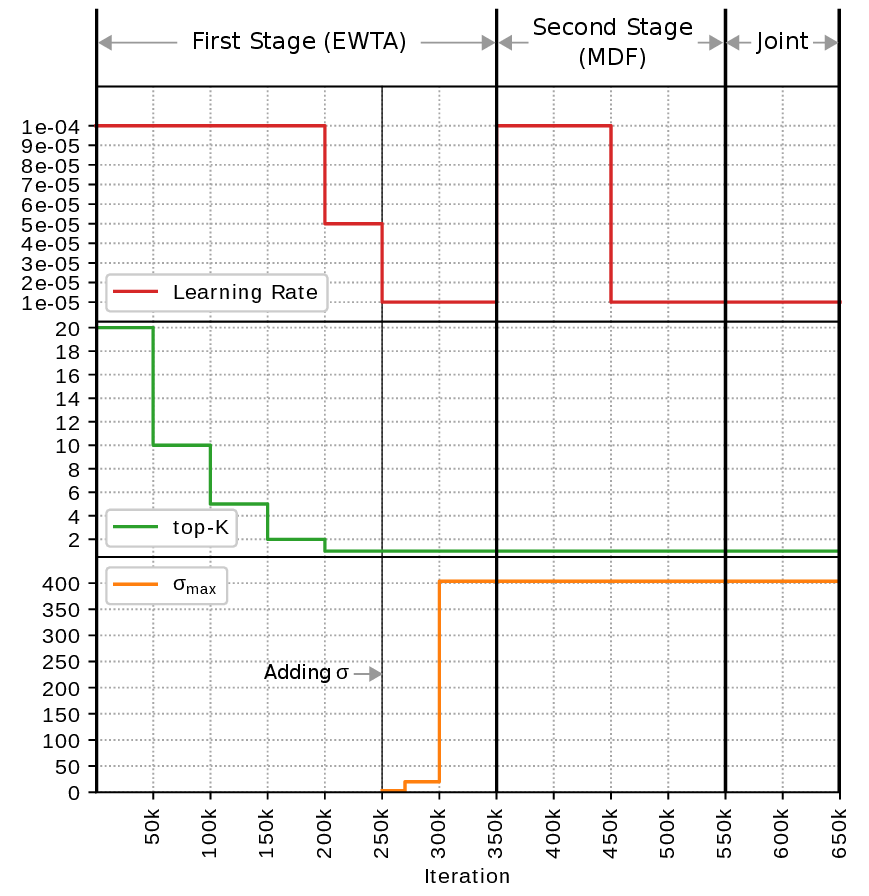}
      \caption{\label{fig:training_two_stage}
      Training schedule for EWTAD-MDF.
      We first train only the first stage for 250k iterations using EWTA, optimizing only the means with $l_{ED}$ (Equation 2 of main paper). At 350k, we add the second stage and train it with the first stage fixed until 550k iterations. After 550k, we remove the loss in the middle and finetune both stages jointly.  
      To stabilize the training we 
      set an upper bound on $\sigma$, which we increase during training. 
      }
      \end{subfigure}
  \end{center}
\caption{
Details of training schedules. 
\label{fig:training_sched}
}
\end{figure*}

Training details for our networks are given in Figure~\ref{fig:training_sched}. To stabilize the training, we also implement an upper bound for $\sigma$ by passing it through a scaled sigmoid function, the slope in the center scaled to $1$.

\section{Ablation Studies}
\subsection{Variants of Sampling-Fitting Framework}
We show a comparison between the two proposed variants of our framework namely EWTAP-MDF and EWTAD-MDF. We observe that the latter leads to better results on both CPI and SDD datasets (see Table~\ref{table:variants_EWTA-MDF}). 
This shows that using WTA with $\l_{NLL}$ (Equation 3 of main paper) and using the predicted uncertainties in the MDF stage is in general better than WTA with $l_{ED}$ (Equation 2 of the main paper). 

\begin{table}
  \begin{center}
  \begin{tabular}{|c|c||c|}
      \hline
      Variant & EMD-CPI & NLL-SDD \\
      \hline
      EWTAP-MDF & $1.70$ & $9.56$\\ 
      EWTAD-MDF & $\textbf{1.57}$ & $\textbf{9.33}$\\ 
      \hline
    \end{tabular}
  \end{center}
  \caption{
  Comparison between the two proposed variants of our sampling-fitting framework. 
  \label{table:variants_EWTA-MDF}
  }
  
\end{table}

\subsection{Effect of History}
We conduct an ablation study on the length of the history for the past $h$ frames. Table~\ref{table:hist} shows the evaluation on both, SDD and CPI. Intuitively, observing longer history into the past improves the accuracy of our proposed framework on CPI. 
However, when testing on SDD, a significant improvement is observed when switching from no history ($h=0$) to one history frame ($h=1$), while only slight difference is observed when using a longer history ($h=2$). This indicates that for SDD only observing one previous frame is sufficient. 
While one past frame allows to estimate velocity, two past frames allow also to estimate also acceleration. This does not seem to be of importance for SDD. 

\begin{table}
  \begin{center}
  \begin{tabular}{|c|c||c|}
      \hline
      $h$ & EMD-CPI & NLL-SDD \\
      \hline
      0.0 & $2.92$ & $15.30$\\ 
      1.0 & $2.13$ & $\textbf{\pz9.13}$\\ 
      2.0 & $\textbf{1.63}$ & $\pz9.30$\\ 
      \hline
    \end{tabular}
  \end{center}
  \caption{
  Evaluation of different lengths of the history used in our EWTAD-MDF. 
  \label{table:hist}
  }
  
\end{table}

\subsection{Effect of Time Horizon}
We conduct an ablation study to analyze the effect of different time horizons in predicting the future. Table~\ref{table:timehorizon} shows the evaluation on both CPI and SDD. Clearly predicting longer into the future is a more complex task and therefore the error increases.

\begin{table}
  \begin{center}
  \begin{tabular}{|c|c||c|c|}
      \hline
      $\Delta t$ (frames) & EMD-CPI & $\Delta t$ (sec) & NLL-SDD \\
      \hline
      10 & $\textbf{1.30}$ & 2.5 & $\pz\textbf{6.94}$\\ 
      20 & $1.74$ & 5   & $\pz8.46$\\ 
      40 & $1.84$ & 10  & $12.21$\\ 
      \hline
    \end{tabular}
  \end{center}
  \caption{
  Evaluation of different time horizons of the future ($\Delta t$) on the proposed framework EWTAD-MDF. 
  \label{table:timehorizon}
  }
  
\end{table}

\subsection{Effect of Number of Hypotheses}
We conduct an ablation study on the number of hypotheses generated by our sampling network EWTAD. Table~\ref{table:numhyps} shows the comparison on CPI and SDD. We observe that generating more hypotheses by the sampling network usually leads to better predictions. However, increasing the number of hypotheses is limited by the capacity of the fitting network to fit a mixture modal distribution, thus explaining the slightly worse results for $K=80$. A deeper and more complex fitting network architecture can be investigated in the future to benefit from more hypotheses.

\begin{table}
  \begin{center}
  \begin{tabular}{|c|c||c|}
      \hline
      $K$ & EMD-CPI & NLL-SDD \\
      \hline
      20 & $1.63$ & $9.33$\\ 
      40 & $\textbf{1.57}$ & $\textbf{9.17}$\\ 
      80 & $1.65$ & $9.22$\\ 
      \hline
    \end{tabular}
  \end{center}
  \caption{
  Evaluation of different number of hypotheses generated by our EWTAD sampling network on the proposed framework EWTAD-MDF. 
  \label{table:numhyps}
  }
\end{table}

\section{Qualitative WTA variant comparison}
\begin{figure*}[t]
  \centering
  \resizebox{1.0\linewidth}{!}{%
  \setlength{\tabcolsep}{0.7pt}%
  \begin{tabular}{ccccc}
      \includegraphics[width=0.2\textwidth, height=1.2in]{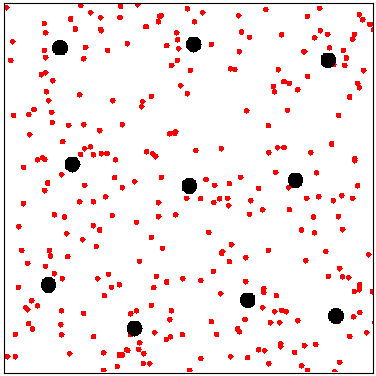}  & &
      \includegraphics[width=0.2\textwidth, height=1.2in]{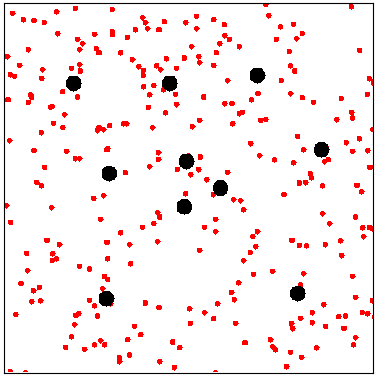} &
      \\
      WTA & & RWTA & &
      \\
      \\
      \includegraphics[width=0.2\textwidth, height=1.2in]{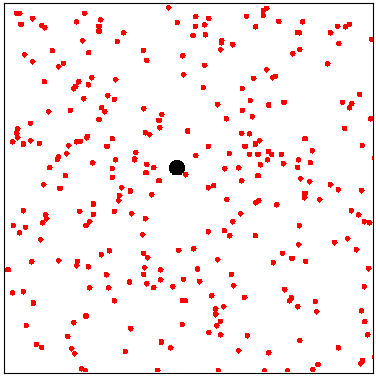} &
      \includegraphics[width=0.2\textwidth, height=1.2in]{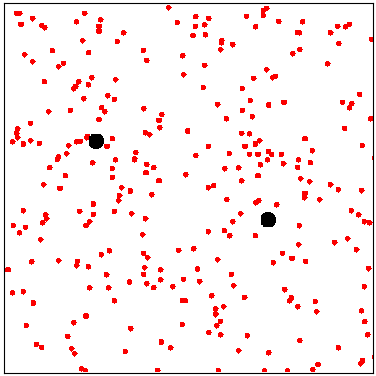} &
      \includegraphics[width=0.2\textwidth, height=1.2in]{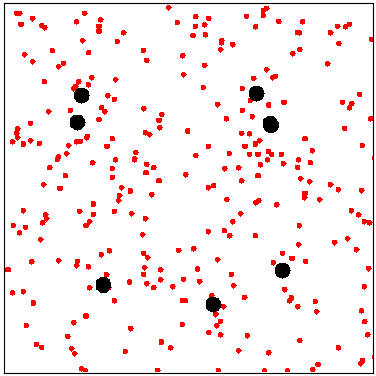} &
      \includegraphics[width=0.2\textwidth, height=1.2in]{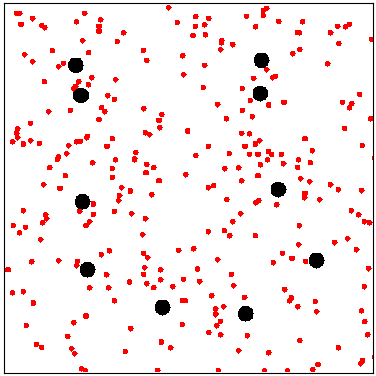} &
      \includegraphics[width=0.2\textwidth, height=1.2in]{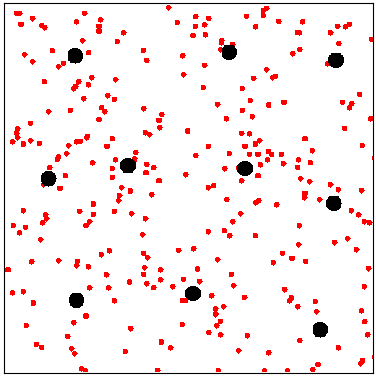}
      \\
      EWTA (Top 10) & EWTA (Top 5) & EWTA (Top 3) & EWTA (Top 2) & EWTA (Top 1)
      \\
      \\
      \includegraphics[width=0.2\textwidth, height=1.2in]{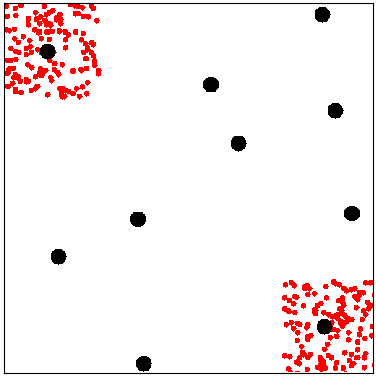}  & &
      \includegraphics[width=0.2\textwidth, height=1.2in]{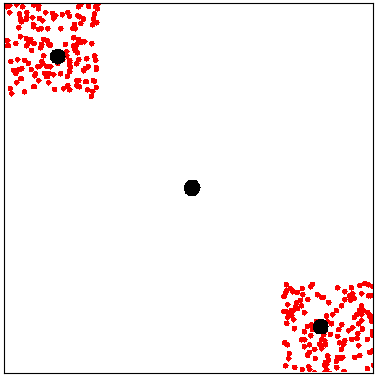} &
      \\
      WTA & & RWTA & &
      \\
      \\
      \includegraphics[width=0.2\textwidth, height=1.2in]{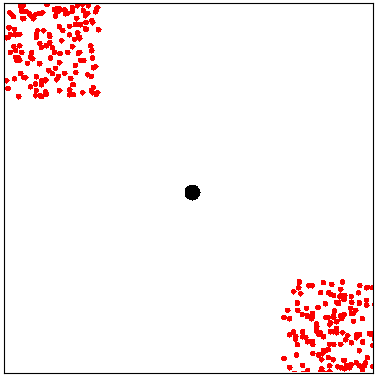} &
      \includegraphics[width=0.2\textwidth, height=1.2in]{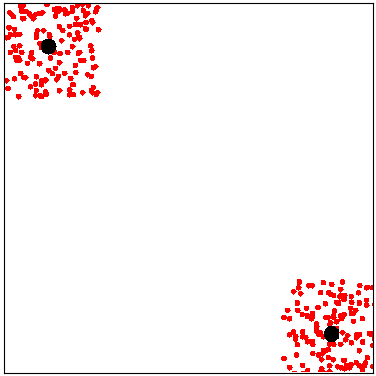} &
      \includegraphics[width=0.2\textwidth, height=1.2in]{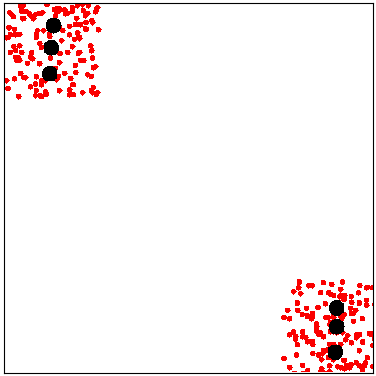} &
      \includegraphics[width=0.2\textwidth, height=1.2in]{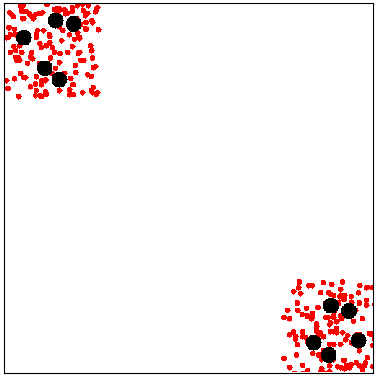} &
      \includegraphics[width=0.2\textwidth, height=1.2in]{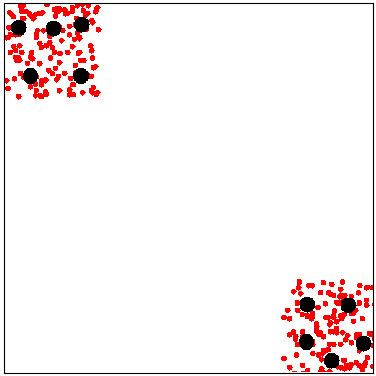}
      \\
      EWTA (Top 10) & EWTA (Top 5) & EWTA (Top 3) & EWTA (Top 2) & EWTA (Top 1)
      \\
\end{tabular}%
}%
   \caption{The simulation results from WTA, RWTA and EWTA. First 2 rows are for uniformly distributed samples over the whole space, while the last 2 rows are uniformly distributed samples centered in upper left and bottom right boxes. 300 ground truth samples are shown as red dots and 10 hypotheses as black dots. EWTA produces hypotheses closer to Voronoi Tessellation. Note that for the third row, 8 hypotheses are moved to the center and only 2 capture the ground-truth samples and RWTA fails to produce a Voronoi Tessellation.
  } 
    \label{fig:wtas}
\end{figure*}

Following~\cite{rupprecht}, we analyze our EWTA in a simulation to see if our variant's hypotheses result in a Voronoi Tesselation. Results are shown in Figure~\ref{fig:wtas}. We see that WTA fails, since it leaves many hypotheses untouched. RWTA similarly leaves 8 hypotheses at the mean position. Our EWTA not only gives hypotheses as close to Voronoi Tesselation as possible, it also assigns equal number of hypotheses to each cluster, which is relevant for distribution fitting.

\section{Failure Cases}
\begin{figure*}[t]
  \centering
  \setlength{\tabcolsep}{0.7pt}%
  \begin{tabular}{ccc}
      \includegraphics[width=0.3\textwidth, height=2.0in]{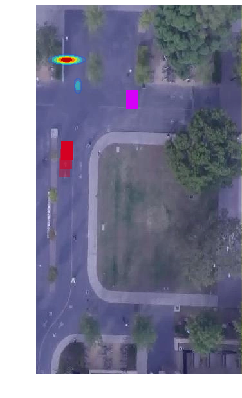}  & 
      \includegraphics[width=0.3\textwidth, height=2.0in]{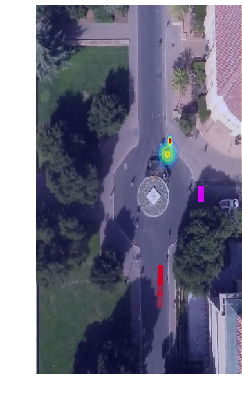}  & 
      \includegraphics[width=0.3\textwidth, height=2.0in]{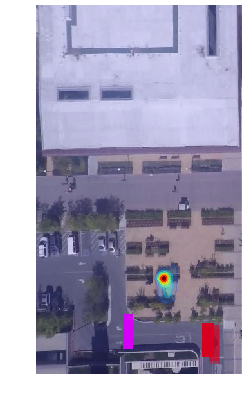}
      \\
      \includegraphics[width=0.3\textwidth, height=2.0in]{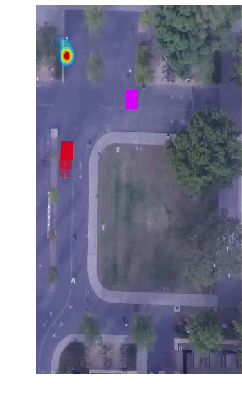} &
      \includegraphics[width=0.3\textwidth, height=2.0in]{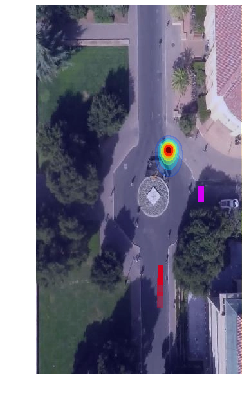} & 
      \includegraphics[width=0.3\textwidth, height=2.0in]{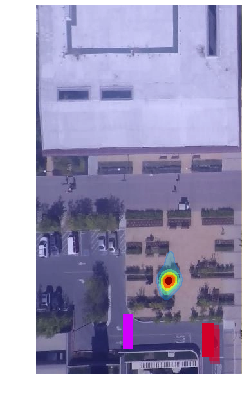}
      \\
\end{tabular}%
   \caption{
   Failure cases for MDN (first row) and our EWTAD-MDF (second row) on SDD. Three past locations of the target object are shown as red boxes, while the ground truth is shown as a magenta box. A heatmap overlay is used to show the predicted distribution over future locations. For interpretation see text.} 
    \label{fig:failure}
\end{figure*}

In Figure~\ref{fig:failure} we depict several failure cases that we found.
We show results for MDN (first row) and our EWTAD-MDF (second row). 
In the first column we see that for a scene that has never been seen during training, both models do not generalize well.
Note that our predicted variance is still more reasonable. 
In the second column, we see another example of missing a mode. This failure is due to the unbalanced training data, where turning right in this scene happens very rarely. 
In the last column, the object of interest is a car, which is an under-sampled class in SDD. The probablity that there is a car in a scene is usually less than $1\%$ and thus this is also a case rarely seen during training. 

\end{document}